\documentclass{winnower}
\usepackage[dvipsnames]{xcolor}
\usepackage{hyperref}
\usepackage{bm}
\usepackage{amsthm}
\usepackage{tikz}
\usepackage[ruled,vlined]{algorithm2e}

\hypersetup{colorlinks,linkcolor={blue},citecolor={orange},urlcolor={red}}  
\makeatother

\newtheorem{claim}{Claim}

\newtheorem{definition}{Definition}
\newtheorem{rem}{Remark}

\newtheorem{property}{Property}

\begin{document}

\title{Nishimori meets Bethe: a spectral method \\for node classification in sparse weighted graphs}

\author[a]{Lorenzo Dall'Amico}
\author[a,b]{Romain Couillet}
\author[a]{Nicolas Tremblay}

\affil[a]{GIPSA-lab Universit\'e Grenoble Alpes, CNRS, Grenoble INP}
\affil[b]{Laboratoire d'Informatique de Grenoble (LIG), Universit\'e Grenoble Alpes}
\affil[$~$]{\texttt{\{lorenzo.dall-amico, romain.couillet, nicolas.tremblay\}@gipsa-lab.fr}
}

\date{\today}

\maketitle

\begin{abstract}
	This article unveils a new relation between the \emph{Nishimori temperature} parametrizing a distribution $P$ and the \emph{Bethe} free energy on random Erd\H{o}s-R\'enyi graphs with edge weights distributed according to $P$. Estimating the Nishimori temperature being a task of major importance in Bayesian inference problems, as a practical corollary of this new relation, a numerical method is proposed to accurately estimate the Nishimori temperature from the eigenvalues of the \emph{Bethe Hessian} matrix of the weighted graph. The algorithm, in turn, is used to propose a new spectral method for node classification in weighted (possibly sparse) graphs. The superiority of the method over competing state-of-the-art approaches is demonstrated both through theoretical arguments and real-world data experiments.
\end{abstract}

\textbf{Keywords}: Clustering techniques, Inference on graphical models, Random matrix theory and extensions, Machine learning.

\section{Introduction}

\subsection{From statistical physics ...}

The physics of disordered systems \citep{binder1986spin} and Bayesian inference for graph learning \citep{jordan1998learning} have long been shown to be tied by a deep connection that has given rise to a host of efficient physics-inspired algorithms \citep{wainwright2008graphical,opper2001advanced, nishimori2001statistical}. A particularly telling example where this relation stands out is the so-called \emph{teacher-student} scenario, in which a set of observed random variables are the outcome of a generative model (the \emph{teacher}) with some hidden parameters to be learned by the \emph{student} \citep{zdeborova2016statistical}.

\medskip

As an instrumental example, we consider in this article the problem of statistical inference on a graph in which the random variable observed by the student is a weighted, undirected graph. Specifically, given a realization of an Erd\H{o}s-R\'enyi graph $\mathcal{G}(\mathcal{V},\mathcal{E})$ with vertex set $\mathcal V$ and edge set $\mathcal E$,\footnote{The graph $\mathcal{G}(\mathcal{V},\mathcal{E})$ is thus \emph{fixed} and not a random variable.} a random \emph{weighted adjacency matrix} $J \in \mathbb{R}^{n \times n}$, with $J_{ij} = J_{ji} \neq 0$  only if $(ij)$ is an edge  of $\mathcal{G}$, is observed by the \emph{student} whose task is to infer some latent variable of the generative model of $J$.
The non-null entries of the matrix $J$ are independently generated by the teacher according to the law
\begin{align}
P(x) = p_0(|x|)e^{\beta_N x},
\label{eq:Nish1}
\end{align}
for an arbitrary non-negative function $p_0(\cdot)$ and some $\beta_{\rm N} > 0$, which we from now on refer to as the \emph{Nishimori temperature}\footnote{For the sake of precision, $\beta_{\rm N}$ behaves as an inverse temperature but, for simplicity, we will refer to it as a \emph{temperature}.} \citep{nishimori1981internal}. The Nishimori temperature naturally appears in statistical physics in the \emph{random bond Ising model} (RBIM), in which the vector $\bm{s} \in \{-1,1\}^n$ is a random variable distributed according to the Boltzmann distribution
\begin{align}
\mu(\bm{s}) = \frac{e^{-\beta\mathcal{H}_J(\bm{s})}}{Z_{J,\beta}},
\label{eq:Boltzmann_distr}
\end{align}
for some positive $\beta$, with $Z_{J,\beta}$ a normalization constant and $\mathcal{H}_J(\bm{s}) = -\bm{s}^TJ\bm{s}$. 

At $\beta = \beta_{\rm N}$, \emph{i.e.}, when the temperature of the system coincides with the Nishimori temperature,\footnote{We underline here that, to be fully rigorous, $\beta_{\rm N}$ is not, by definition, a temperature, but rather a parameter of the generative model of $J$, \emph{i.e.}, a hidden parameter of the teacher's generative model.} the exact expression of $\mathbb{E}\left[\langle \mathcal{H}_{J}(\bm{s})\rangle_{\beta}\right]$ can be computed with elementary mathematical tools, where $\langle \cdot \rangle_{\beta}$ denotes the averaging over the Boltzmann distribution \eqref{eq:Boltzmann_distr} while $\mathbb{E}[\cdot]$ is the averaging over the realizations of $J$ distributed as \eqref{eq:Nish1}. It has also been shown \citep{nishimori2001absence, zdeborova2016statistical} that the RBIM at the Nishimori temperature is either in the \emph{ferromagnetic} configuration (in which $\langle s_i \rangle_{\beta} > 0$ for all $i$) or in the \emph{paramagnetic} configuration (for which $\langle s_i \rangle_{\beta} = 0$ for all $i$). In particular, the system is never in the \emph{spin-glass phase} under which local order of $\bm s$ appears despite there being no global magnetization. These relevant properties drew research attention to this particular temperature \citep{georges1985exact, gruzberg2001random, toldin2009strong} since its first appearance in \citep{nishimori1981internal}.

\subsection{... to Bayesian inference on weighted sparse networks}

The importance of the \emph{Nishimori temperature} in Bayesian inference was thoroughly discussed in \citep{iba1999nishimori}, where the author exhibits a correspondence between the optimal Bayes inference problem (\emph{i.e.}, when the \emph{student} knows exactly the generative model of the \emph{teacher}) and the RBIM studied at $\beta_{\rm N}$. 

\medskip 

As a practical and telling example of modern concern of the importance of the Nishimori temperature in Bayesian statistics, we here consider as a common thread the problem of binary node classification on a graph. Specifically, let $\bm{\sigma} \in \{-1,1\}^n$ be a label vector assigning each node to a ``class''. Further assume that a matrix $J$ is drawn from the distribution \eqref{eq:Nish1} and that the student has to infer the vector $\bm{\sigma}$ from the observation of the matrix $\tilde{J}$, defined by $\tilde{J}_{ij} = J_{ij}\sigma_i\sigma_j$. The matrix $\tilde{J}$ has entries that, \emph{in expectation}, are positive if nodes $i$ and $j$ have the same label and negative otherwise. As discussed extensively in Section~\ref{sec:ml}, this quite elementary model can in fact be used to study \emph{correlation clustering} over the $p$-dimensional \emph{feature} vectors $\bm z_1,\ldots,\bm z_n\in\mathbb R^p$ of a dataset of size $n$ \citep{bansal2004correlation}, with concrete application to image, sound, or sentence classification \citep{langone2016kernel}. In this example, the weights $J_{ij}$ carried by the edges of $\mathcal G$ represent some affinity metric between the features $\bm z_i$ and $\bm z_j$ associated with nodes $i$ and $j$ (in essence, the larger $J_{ij}$ the closer $\bm z_i$ and $\bm z_j$).

\medskip

From a Bayesian perspective, inferring $\bm{\sigma}$ from $\tilde{J}$ reduces to computing the marginals of the distribution
\begin{align}
\mathbb{P}(\bm{\sigma}|\tilde{J}) = \frac{e^{-\beta_{\rm N} \mathcal{H}_{\tilde{J}}(\bm{\sigma})}}{Z_{\tilde J,\beta_{\rm N}}} .
\label{eq:Bayes_opt}
\end{align}
This thus coincides with computing the \emph{magnetizations} $\bm{m} = \langle \bm{\sigma} \rangle_{\beta_{\rm N}}$  of a RBIM on the graph $\tilde{J}$ at the Nishimori temperature.
However, assuming that the observing \emph{student} knows the value of $\beta_{\rm N}$ is often unrealistic (in effect, the student only sees $\tilde J$) and earlier works have resorted to studying the problem of \emph{mismatched inference} (\emph{i.e.}, inference performed when the \emph{student} uses a different parameter than the one assumed by the \emph{teacher}) \citep{zdeborova2016statistical}. 

\subsection{Our contribution: relating Nishimori to Bethe}

Our main result consists in going beyond mismatched inference by providing an efficient estimate to the Nishimori temperature.
To this end, we first draw an explicit relation between the Nishimori temperature and the smallest eigenvalue of the Hessian matrix of the Bethe free energy associated to the RBIM, when set at the paramagnetic point $\bm{m} = \bm{0}$ (this Hessian matrix is the so-called \emph{Bethe-Hessian matrix} \citep{watanabe2009graph}); this relation holds under the previously introduced setting, so in particular for a student observation matrix $J$ supported over a (possibly sparse) Erd\H{o}s-R\'enyi graph $\mathcal G$. Besides, we observe and argue that, although the Bethe approximation is particularly adapted to sparse (locally tree-like) graphs $\mathcal G$, the \emph{Nishimori-Bethe relation} holds for any degree of sparsity (that is, even when $\mathcal G$ does not behave locally as a tree). 

\medskip 

The main consequences of the Nishimori-Bethe relation, and our main contributions, consist in
\begin{itemize}
    \item[i)] the design of a new efficient spectral algorithm which estimates the Nishimori temperature with asymptotically perfect accuracy (as $n\to\infty$); the algorithm is based on an iterative fast search of a well-parametrized Bethe-Hessian matrix exhibiting a smallest amplitude eigenvalue close to zero;
    \item[ii)] a new spectral algorithm to approximately solve the Bayesian node classification inference problem of Equation~\eqref{eq:Bayes_opt}, which outperforms commonly deployed state-of-the-art alternatives. We in particular claim that this spectral algorithm is capable of performing non trivial inference as soon as the Bayesian optimal solution can;
    \item[iii)] although we claim that these algorithms are still valid under dense graphs $\mathcal G$, they are specifically adapted to the sparse regime where $|\mathcal{V}| \sim |\mathcal{E}|$; this practically allows for small computational and memory storage costs when applied to the classification of the nodes of possibly large graphs; we specifically support this fact by a concrete application to the classification of $40\,000$ high resolution images using our proposed sparse but extremely efficient spectral algorithm.
\end{itemize}

\medskip

The remainder of the article is structured as follows. Section~\ref{sec:basic} introduces the RBIM together with some basic properties of the Nishimori temperature. These serve as the support for Section~\ref{sec:main}, which provides our main results: the Nishimori-Bethe relation, the aforementioned new algorithms to estimate the Nishimori temperature, and how it provides an approximate (but still accurate) solution to the Bayesian inference problem. To corroborate the claims made in this section, Section~\ref{sec:ml} applies the results to a concrete node classification problem involving realistic images produced by generative adversarial networks \citep{brock2018large}.  Section~\ref{sec:conclusion} closes the article laying out some limitations and possible directions of improvement of the present analysis.

\medskip

A Julia implementation of our proposed algorithm as well as the codes used to produce the results of this article is available at \href{https://github.com/lorenzodallamico/NishimoriBetheHessian}{github.com/lorenzodallamico/NishimoriBetheHessian}.

\medskip

\textbf{Notation}: Vectors are denoted in bold face. The notation $\bm{1}_n$ indicates the all ones vector of size $n$. Scalar and matrices are in standard font, with matrices denoted by capital letters. The notation $`\circ`$ indicates the Hadamard entry-wise product between two matrices of same size. The notation $\partial i$ indicates the neighborhood of node $i$ on the graph $\mathcal{G}(\mathcal{V},\mathcal{E})$: $\partial i = \{j \in \mathcal{V}~:~(ij) \in\ \mathcal{E}\}$.

\section{Basic properties of the random bond Ising model}
\label{sec:basic}

In this section we provide the basic language and properties necessary to define the Nishimori temperature. The results presented in this section do not all have a direct application to inference problems, which are discussed later in Section~\ref{sec:ml}.

\subsection{Phase diagram}

Consider a realization of a Erd\H{o}s-R\'enyi graph $\mathcal{G}(\mathcal{V},\mathcal{E})$ with expected average degree $c$. We will denote $\mathcal{V}$ the set of the $n$ nodes of the graph and $\mathcal{E}$ the set of its edges. We further let $J \in \mathbb{R}^{n \times n}$ be a weighted adjacency matrix on $\mathcal{G}$ and distributed according to the following generative model. 

\begin{definition}[Generative model of $J$]
	\label{def:Nish}
	For all edges $(ij) \in \mathcal{G}$ with $i > j$, the $J_{ij}$ are generated independently (with $J_{ij} = J_{ji}$), for some $\beta_{\rm N} > 0$, referred to as \emph{Nishimori temperature}, according to  
	\begin{align}
	\forall~(ij) \in \mathcal{E},~i<j&,\quad J_{ij} \overset{\rm i.i.d.}{\sim} P \nonumber\\ 
	P(x) &= p_0(|x|)~e^{\beta_{\rm N}x},
	\label{eq:Nish_for_J}
	\end{align}	
	where $p_0(\cdot)$ is an arbitrary non-negative function satisfying the normalization condition $\int_{-\infty}^{\infty}dx~ p_0(|x|)e^{\beta_{\rm N}x}~=~1$.
	If $(ij) \notin \mathcal{E}$, then $J_{ij}=0$.
\end{definition}

Given a realization of $J$ and a vector $\bm{s} \in \{-1,1\}^n$, we define the \emph{Hamiltonian} $\mathcal{H}_J(\bm{s})$ of the RBIM as
\begin{align}
\mathcal{H}_{J}(\bm{s}) = -\sum_{(ij)\in\mathcal{E}} J_{ij}s_is_j = -\bm{s}^TJ\bm{s}.
\label{eq:ham}
\end{align}

\begin{figure}[t!]
	\centering
	\includegraphics[width=0.5\columnwidth]{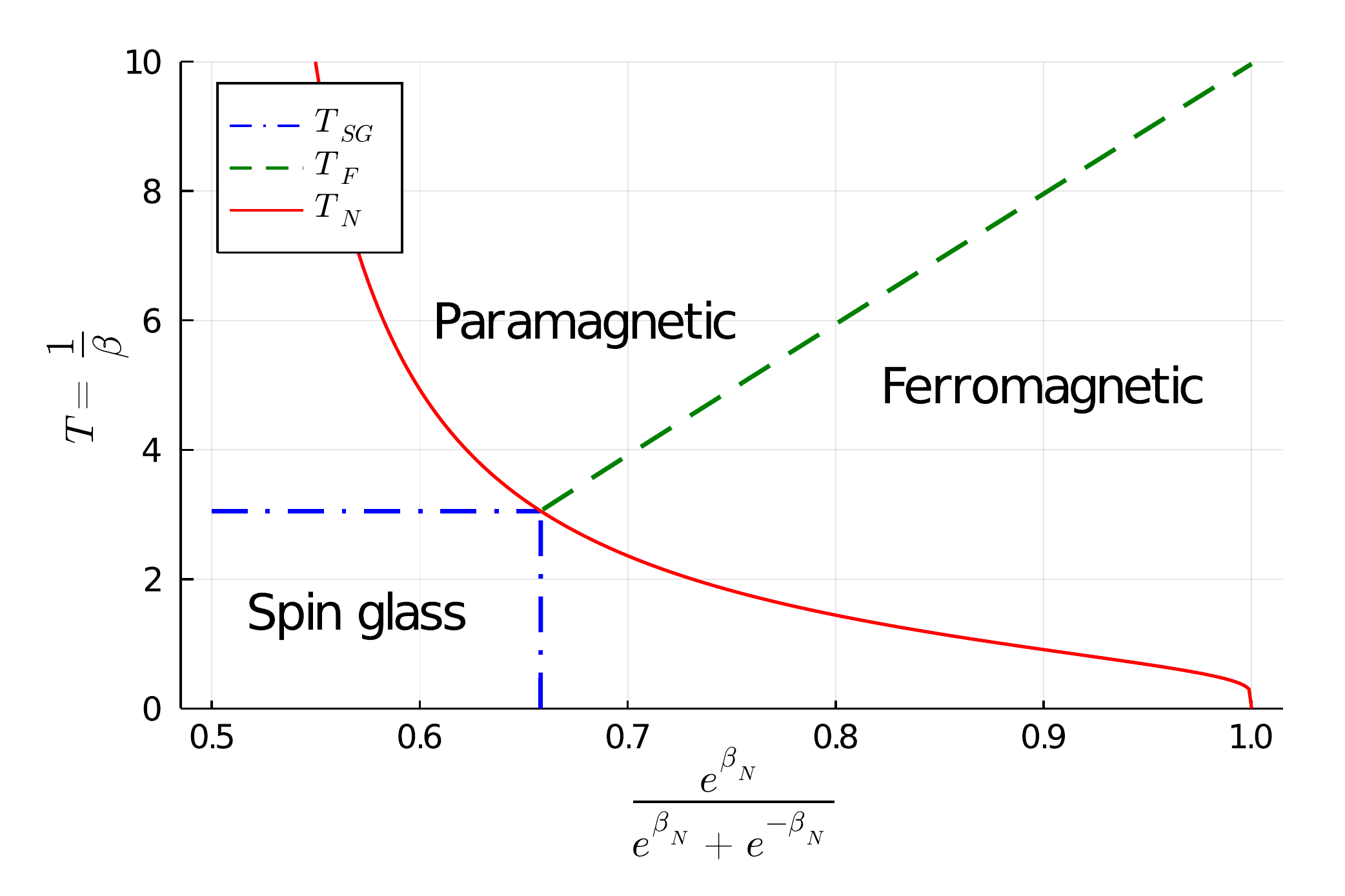}
	\caption{Phase diagram of the RBIM for $J_{ij } \in \{-1,1\}$. The $x$ axis goes from $\frac{1}{2}$ for $\beta_{\rm N} = 0$ to $1$ for $\beta_{\rm N} \to \infty$. The $y$ axis represents $T$, the inverse of $\beta$. The dashed green line is the inverse of $\beta_{\rm F}$, the dash dotted blue line is the inverse of $\beta_{\rm SG}$ and the solid red line is the inverse of $\beta_{\rm N}$.}
	\label{fig:ph_d}
\end{figure}

Note that, from Definition~\ref{def:Nish}, the Nishimori temperature is defined \emph{independently} of $\mathcal{G}$, while the dependence of $p_0(\cdot)$ on $\beta_{\rm N}$ is relegated to its normalization constant. Two examples of distributions that fall under this definition are the $\pm J$ model
\begin{align*}
P(x) =  p\delta(x - J_0) + (1-p)\delta(x + J_0), \quad {\rm for}~p\in[1/2,1], \quad J_0 \in \mathbb{R}^+ 
\end{align*}
that can be rewritten as
\begin{align*}
    P(x) &= \frac{e^{\beta_{\rm N}x}}{2{\rm ch}(\beta_{\rm N}J_0)}, \quad {\rm with} \quad
    \beta_{\rm N} = \frac{1}{2J_0}~{\rm log}\frac{p}{1-p},
\end{align*}
and the Edwards-Andersons model \citep{edwards1975theory}
\begin{align*}
P(x) &= \frac{1}{\sqrt{2\pi\nu^2}}~{\rm exp}\left\{-\frac{(x-J_0)^2}{2\nu^2}\right\}, \quad {\rm for}~ J_0,\nu \in \mathbb{R}^+ 
\end{align*}
for which
\begin{align*}
    p_0(|x|) &= \frac{1}{\sqrt{2\pi\nu^2}}~{\rm exp}\left\{-\left(\frac{x^2+J_0^2}{2\nu^2}\right)\right\} \\
    \beta_{\rm N} &= \frac{J_0}{\nu^2}.
\end{align*}
Given a matrix $J$ drawn from the generative model of  Definition~\ref{def:Nish}, Equation~\eqref{eq:ham} and a temperature $\beta \in \mathbb{R}^+$, we now let $\bm{s} \in \{-1,1\}^n$ be a random vector, drawn from the Boltzmann distribution
\begin{equation}
\mu(\bm{s}) = \frac{e^{-\beta \mathcal{H}_{J}(\bm{s})}}{Z_{J,\beta}},
\label{eq:Boltzmann}
\end{equation}
where $Z_{J,\beta}$ is the normalization constant. Averaging over the distribution~\eqref{eq:Boltzmann} will be denoted with $\langle \cdot \rangle_{\beta}$.

\medskip

Let us now consider the phase diagram, depicted in Figure~\ref{fig:ph_d}, of the model described by Equations~(\ref{eq:ham}, \ref{eq:Boltzmann}) and Definition~\ref{def:Nish}. First consider the role played by the two parameters $\beta$ and $\beta_{\rm N}$. For increasing values of $\beta_{\rm N}$, there is a larger probability for each edge $J_{ij}$ to carry a positive weight and the minimum of $\mathcal{H}_J(\bm{s})$ is achieved for $\bm{s} = \bm{1}_n$. For small values of $\beta_{\rm N}$, instead, multiple local minima appear. Concerning $\beta$, instead, for small values, the Boltzmann distribution (Equation~\eqref{eq:Boltzmann}) tends towards a uniform distribution, while, for large values, the configurations with small energy $\mathcal{H}_J(\bm{s})$ have a larger probability. Consequently, for large $\beta$ and $\beta_{\rm N}$ the average configuration of $\bm{s}$ tends to align towards $\bm{1}_n$: this corresponds to the \emph{ferromagnetic} configuration. Conversely, for small values of $\beta_{\rm N}$, several edges carry a negative weight, introducing \emph{frustration} in the system that is found in the \emph{spin-glass} phase, for which local order of the spins may be observed ($\frac{1}{n}\sum_i \langle s_i \rangle_{\beta}^2 \neq 0$), but globally the magnetization is null ($\frac{1}{n} \sum_i \langle s_i \rangle_{\beta} = 0$). Finally, at large values of $\beta$, the system is in the \emph{paramagnetic} phase, for which the spins are randomly aligned and the magnetization is zero.

\medskip 

In the particular case where $\mathcal{G}$ is an Erd\H{o}s-R\'enyi random graph, with expected average degree equal to $c$, the cavity method \citep{mezard2009information} predicts the position of the transitions between the three phases: the \emph{paramagnetic-ferromagnetic} transition occurs at $\beta = \beta_{\rm F}$ and the  \emph{paramagnetic-spin glass} transition occurs at $\beta = \beta_{\rm SG}$, also known as the de Almeida-Thouless transition \citep{thouless1986spin}. The values of $\beta_{\rm F}, \beta_{\rm SG}$ are given as the solutions of the following equations \citep{zdeborova2016statistical} :
\begin{align}
&c\cdot \mathbb{E}[{\rm th}(\beta_{\rm F}J_{ij})] := 1 \label{eq:def_transitions_F}\\
&c\cdot \mathbb{E}[{\rm th}^2(\beta_{\rm SG}J_{ij})] := 1,
\label{eq:def_transitions_SG}
\end{align}
where we recall that $\mathbb{E}[\cdot]$ denotes averaging over the distribution \eqref{eq:Nish_for_J}.
Figure~\ref{fig:ph_d} precisely depicts the phase diagram for the $\pm J$ model. A qualitatively similar diagram can be obtained for different distributions, that follow the definition of Equation~\eqref{eq:Nish1} \citep{nishimori1981internal}. Given these premises, we now discuss some relevant properties valid on the Nishimori line, \emph{i.e.} when $\beta = \beta_{\rm N}$.

\subsection{Relevant properties at the Nishimori temperature}

First of all, let us introduce the \emph{quenched internal energy density}, defined as $u(\beta) := \frac{1}{n} \mathbb{E}[\langle\mathcal{H}_J(\bm{s})\rangle_{\beta}]$, where we recall that $\langle \cdot \rangle_{\beta}$ denotes an average taken over the Boltzmann distribution~\eqref{eq:Boltzmann} and $\mathbb{E}[\cdot]$ is the average over the distribution of Equation~\eqref{eq:Nish1}. It was shown in \citep{nishimori1981internal} that $u(\beta_{\rm N})$ takes a particularly simple expression:
\begin{align*}
u(\beta_{\rm N}) =\frac{1}{n} \mathbb{E}[\langle\mathcal{H}_J(\bm{s})\rangle_{\beta_{\rm N}}] = - \frac{1}{n}\sum_{(ij)\in\mathcal{E}}\mathbb{E}\left[ J_{ij}\langle s_is_j\rangle_{\beta_{\rm N}}\right] = -\frac{1}{n}\sum_{(ij)\in\mathcal{E}}\mathbb{E}[J_{ij}~{\rm th}(\beta_{\rm N}J_{ij})].
\end{align*}

The first two equalities are true by definition. The elegance of the result of \citep{nishimori1981internal} lies in the last relation that identifies -- inside the expectation $\mathbb{E}[\cdot]$ -- the term $\langle s_is_j \rangle_{\beta_{\rm N}}$ with ${\rm th}(\beta_N J_{ij})$. We will show in Section~\ref{sec:main.phd} that, for $\beta$ sufficiently small, the system is in the paramagnetic phase $\langle \bm{s} \rangle_{\beta} = 0$ and, under the Bethe approximation, the relation $\langle s_is_j \rangle_{\beta} = {\rm th}(\beta J_{ij})$ is verified for any underlying $\beta_{\rm N}$. This informally introduces a relation between the Bethe free energy at the paramagnetic point and the Nishimori temperature, which is at the centre of Claim~\ref{conj:1}.

\medskip

We introduce the following property of the probability distribution of Equation~\eqref{eq:Nish1}. This relation will be of fundamental use in the following and, in passing, allows us to rewrite $u(\beta_{\rm N})$ as in \citep{nishimori1981internal}.

\begin{property}
\label{prop:th}
	Let $f(x)$ be an arbitrary odd function. Then
	\begin{align}
	\mathbb{E}[f(x)\cdot{\rm th}(\beta_{\rm N}x)] = \mathbb{E}[f(x)].
	\end{align}
\end{property}
The proof is easily obtained by straightforward calculation.
As a consequence of Property~\ref{prop:th}, the quenched internal energy density at the Nishimori temperature takes the simple expression:
\begin{align*}
u(\beta_{\rm N}) = -\frac{1}{n}\sum_{(ij)\in\mathcal{E}} \mathbb{E}[J_{ij}] = -\frac{\bar{d}}{2}\cdot\mathbb{E}[J_{ij}],
\end{align*}
where $\bar{d}$ denotes the average node degree in the graph $\mathcal G$. 

\medskip

Secondly, we recall a well celebrated property of the Nishimori temperature, which states the absence of \emph{replica symmetry breaking} on the Nishimori line \citep{nishimori2001absence, zdeborova2016statistical} or, equivalently, that the RBIM at the Nishimori temperature is never in the spin glass phase.
This result can be visually understood in Figure~\ref{fig:ph_d} by noticing that the Nishimori temperature is either in the paramagnetic or ferromagnetic phase. Moreover, exploiting Property~\ref{prop:th} and the definitions of $\beta_{\rm F}, \beta_{\rm SG}$ in Equations~(\ref{eq:def_transitions_F}, \ref{eq:def_transitions_SG}), on an Erd\H{o}s-R\'enyi graph one finds that $\beta_{\rm SG} = \beta_{\rm N} \iff \beta_{\rm F} = \beta_{\rm N}$. Consequently, there exists a tricritical point where $\beta_{\rm F} = \beta_{\rm SG} = \beta_{\rm N}$. 

\medskip

Recalling the connection with statistical inference problems, such as inferring $\bm\sigma$ in Equation~\eqref{eq:Bayes_opt}, first note that $\beta_{\rm N}$ is the Bayes optimal inference temperature in the sense that there exists no other $\beta$ that can asymptotically achieve better inference performance and, therefore, if inference cannot be performed at $\beta = \beta_{\rm N}$, then it is theoretically infeasible. This occurs when the marginals of Equation~\eqref{eq:Bayes_opt} asymptotically give equal probabilities for each $\sigma_i$ to take either values $\pm 1$. In terms of the phase diagram, this corresponds to being in the \emph{paramagnetic} phase, so that $\beta_{\rm N} < \beta_{\rm F}$. In order for non-trivial reconstruction to be possible, the condition $\beta_{\rm N} < \beta_{\rm SG} < \beta_{\rm F}$ must be imposed \citep{saade2016clustering}. When the condition is met, the system is in the \emph{informative} configuration in which each \emph{spin} gets oriented towards its planted value $\sigma_i$. This being said, replacing (or effectively, erroneously estimating) $\beta_{\rm N}$ by $\beta \neq \beta_{\rm N}$ in Equation~\eqref{eq:Bayes_opt}, it may occur that, even though inference is theoretically possible (as $\beta_{\rm N} < \beta_{\rm SG} < \beta_{\rm F}$), the estimated labels $\hat{\bm{\sigma}}$ for the mismatched $\beta$ are not aligned with the ground truth $\bm{\sigma}$. This never happens at $\beta = \beta_{\rm N}$ for which inference is achieved \emph{as soon as theoretically possible}.

\medskip

With this short introduction on the Nishimori temperature at hand, in the next section we present our main result which relates $\beta_{\rm N}$ to the spectrum of the non-backtracking and Bethe-Hessian matrices of the underlying graph $\mathcal G$.

\section{A relation between $\beta_{\rm N}$ and the Bethe free energy}
\label{sec:main}

This section introduces our main theoretical result, which draws a connection between the Nishimori temperature and the \emph{variational free energy under the Bethe approximation}, computed at the \emph{paramagnetic} point $\langle \bm{s}\rangle_{\beta} := \bm{m} = 0$. To this end, Section~\ref{sec:main.preli} introduces two fundamental matrices, namely the \emph{non-backtracking} and the \emph{Bethe-Hessian} matrices of the graph $\mathcal G$ and recalls the known connections between the spectra of these two matrices. Section~\ref{sec:main.main} then introduces our main result, precisely consisting in (i) a claim on the location of the eigenvalues of the \emph{non-backtracking} matrix and, as a result of the claim, (ii) an explicit relation between the underlying Nishimori temperature and a specific eigenvalue of the \emph{non-backtracking} matrix. We further provide both theoretical arguments and numerical simulations in support of the result. As a corollary of the identities listed in Section~\ref{sec:main.preli} and Section~\ref{sec:main.main}, we finally obtain an explicit relation between the \emph{smallest eigenvalue of the Bethe-Hessian matrix} and the Nishimori temperature. Section~\ref{sec:main.phd} relates this central link to the phase diagram of Figure~\ref{fig:ph_d}, in passing connecting the results to the expression of the Bethe free energy. Based on these findings, Section~\ref{sec:main.algo} elaborates an algorithm to estimate $\beta_{\rm N}$, which finds significant importance in statistical inference problems.

\subsection{Preliminaries}
\label{sec:main.preli}

Let us first introduce the weighted \emph{non-backtracking} matrix of any arbitrary graph $\mathcal G$.

\begin{definition}[Weighted non-backtracking matrix]
Given a graph $\mathcal{G}(\mathcal{V},\mathcal{E})$ and a function $f~:~ \mathcal{E}~\to~\mathbb{R}$, so that $\forall~ e \in \mathcal{E}$, $f(e) = \omega_e$ is the
weight corresponding to the edge $e$, the weighted non backtracking matrix $B \in \mathbb{R}^{2|\mathcal{E}|\times 2|\mathcal{E}|}$ is defined on the set
of directed edges of $\mathcal{G}$ as
\begin{align}
B_{(ij),(k\ell)} = \delta_{jk}(1-\delta_{i\ell})\omega_{k\ell}.
\label{eq:B}
\end{align}
\label{def:B}
\end{definition}

The \emph{non-backtracking} matrix plays an important role in inference and graph mining problems \citep{krzakala2013spectral,zhang2015nonbacktracking,aleja2019non, torres2019non,torres2020node,arrigo2020beyond, shi2018weighted} and naturally comes into play from the linearization of the \emph{belief propagation} (or \emph{cavity}) equations \citep{mezard2009information} for the RBIM. These equations are particularly adapted to dealing with locally \emph{tree-like} structured graphs (such as sparse Erd\H{o}s-R\'enyi graphs).

\medskip

The eigenvalues of the matrix $B$ are strongly related to the eigenvalues of the \emph{Bethe-Hessian} matrix.

\begin{definition}[Bethe-Hessian matrix]
	\label{def:BH}
	Given a graph $\mathcal{G}(\mathcal{V},\mathcal{E})$, a function $f:\mathcal{E} \to \mathbb{R}$ so that $\forall~e\in\mathcal{E}$, $f(e) = \omega_e$ and a parameter $x \in \mathbb{C}\setminus \{\pm \omega_{ij}\}_{(ij)\in\mathcal{E}}$, the Bethe-Hessian matrix $H(x) \in \mathbb{C}^{n\times n}$ is defined as
	\begin{align}
	H_{ij}(x) = \left(1+\sum_{k \in \partial i} \frac{ \omega^2_{ik}}{x^2-\omega_{ik}^2}\right)\delta_{ij} - \frac{x~\omega_{ij}}{x^2-\omega^2_{ij}}.
	\label{eq:BH}
	\end{align}
\end{definition}
Since $\mathcal{G}$ is an undirected graph, $H(x)$ is symmetric but not Hermitian, unless $x \in \mathbb{R}$.
The relation between the spectra of the matrices $B$ and $H(x)$ is given by the Watanabe-Fukumizu formula \citep{watanabe2011loopy,sato2014matrix}.

\begin{property}[Watanabe-Fukumizu]
Let $H(x)$ and $B$ be defined as per (\ref{eq:B}, \ref{eq:BH}) on the same graph $\mathcal{G}$ and for the same weighting function $f$. Further let $x\in\mathbb{C}\setminus \{\pm \omega_{ij}\}_{(ij)\in\mathcal{E}}$. Then,
\begin{align}
{\rm det}\big[xI_{2|\mathcal{E}|} - B\big] = {\rm det}\big[H(x)\big] \prod_{(ij)\in\mathcal{E}}  \big( x^2 -\omega_{ij}^2\big),
\end{align}
so that, for all $x$ in the spectrum of $B$, ${\rm det}[H(x)]=0$.
\label{prop:ihara}
\end{property}

With this preliminary information, we now proceed to the formulation of our main result which first consists in a conjecture on the spectrum of $B$, and which we then relate to the spectrum of $H(x)$ through Property~\ref{prop:ihara}. Choosing $f(e) = {\rm th}(\beta J_e)$ in the definition of $B$, where the weights $J_e$ are distributed according to Equation~\eqref{eq:Nish_for_J}, we finally unfold the relation between the spectra of $B$, $H(x)$ and the Nishimori temperature.

\subsection{Main result}
\label{sec:main.main}

We now proceed to studying the spectrum of the matrix $B$ in the case where $\mathcal{G}$ is a Erd\H{o}s-R\'enyi graph and its weights $\omega_e$ (Equation~\eqref{eq:B}) are drawn i.i.d.\@ satisfying $|\omega_e| < 1$ with $\mathbb{E}[\omega] > 0$ sufficiently large. The interest of this setting in relation to the RBIM and the Nishimori temperature is to consider $\omega_e = {\rm th}(\beta J_e)$ for $\beta_{\rm N} > \beta_{\rm SG}$ and $J$ as per Definition~\ref{def:Nish}. In this particular case, one of the eigenvalues  of $B$ -- and, as a consequence of Property~\ref{prop:ihara}, a corresponding (more easily estimated) eigenvalue of the Bethe-Hessian matrix -- has a direct relation with $\beta_{\rm N}$.

\medskip

\begin{figure}
	\centering
	\includegraphics[width=0.7\columnwidth]{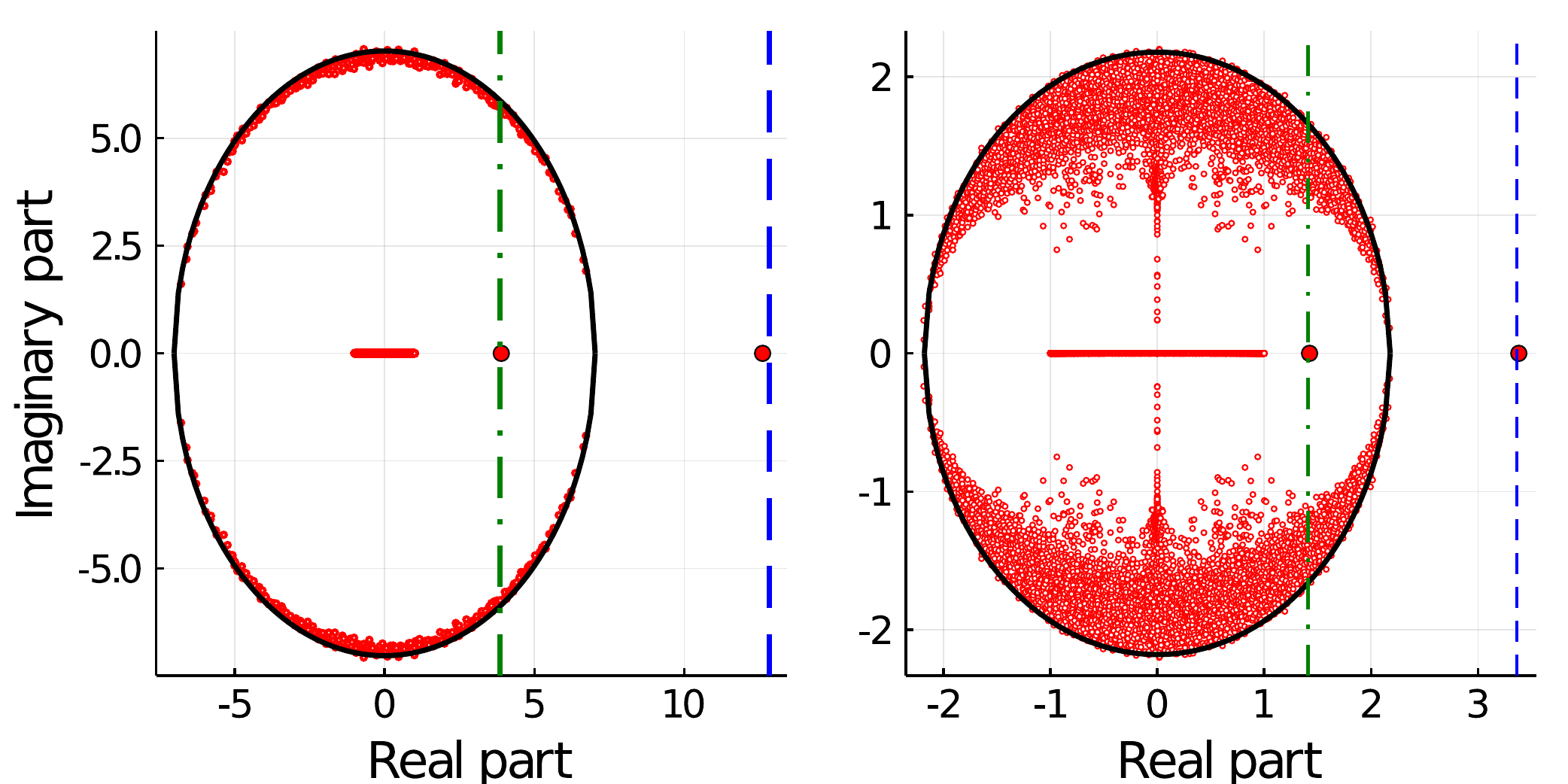}
	\caption{Spectrum of the matrix $B$ in the complex plane. The entries $J_{ij}$ are generated independently  according to  $\mathcal N(J_0,\nu^2)$. The weights appearing in Equation~\eqref{eq:B} are defined as $\omega_{ij} = {\rm th}(\beta J_{ij})$. \textbf{Left, dense regime}: $n = 250$, $c = 2~{\rm log}^2(n) $, $J_0 = 1$, $\nu = 4$, $\beta = 1$. \textbf{Right, sparse regime}: $n = 3\,000$, $c = 5$, $J_0 = 1$, $\nu = 1$, 	$\beta = 10$. \textbf{For both plots}, the dashed blue line corresponds to $ c\mathbb{E}[{\rm th}(\beta J)]$, the dash-dotted green line  to $\mathbb{E}[{\rm th}^2(\beta J)]/\mathbb{E}[{\rm th}(\beta J)]$, while the black continuous line is the circle in the complex plane centered at the origin and of radius $\sqrt{c\mathbb{E}[{\rm th}^2(\beta J)]}$.}
	\label{fig:spec_B}
\end{figure}

The matrix $B$ is not symmetric, hence its eigenvalues are not necessarily real. Since $B$ is real though, the non-real eigenvalues come in complex-conjugate pairs.
When weights are assigned independently at random in the interval $(-1,1)$, we observe, in agreement with the theoretical results obtained on the spectrum of $B$ \citep{gulikers2016non,bordenave2015non,stephan2020non,coste2019eigenvalues}, that in the $n\to \infty$ limit, the non-real eigenvalues of $B$ are bounded by a circle on the complex plane and are separated by a vanishing distance from one another. These eigenvalues form the \emph{bulk} of the spectrum of $B$ (see Figure~\ref{fig:spec_B}). There further exists one \emph{real} eigenvalue which is instead \emph{isolated}, \emph{i.e.}, it is found at a macroscopic (not decreasing with $n$) distance from all other eigenvalues. This eigenvalue has a modulus greater than the radius of the bulk: its existence and position are known and have been thoroughly investigated \citep{stephan2020non,coste2019eigenvalues}. There however exists another real isolated eigenvalue with modulus \emph{smaller} than the radius of the bulk, the existence and importance of which were first evidenced in \citep{dall2019revisiting} in the case of unweighted graphs with a community structure. After \citep{dall2019revisiting}, a similar phenomenon has also been observed in \citep{maillard2020construction} in the context of phase retrieval, relating the Hessian of the TAP free energy and the Bayes optimal inference temperature. This isolated eigenvalue inside the bulk of $B$ received less theoretical attention and it is the main object of our central result.

\begin{claim}
	Let $\mathcal{G}(\mathcal{V},\mathcal{E})$ be a realization of an Erd\H{o}s-R\'enyi random graph with $n$ nodes ($n \to \infty$) and expected average degree $c$. For each undirected edge $(ij) \in \mathcal{E}$ a weight $\omega_{ij}=\omega_{ji} \in(-1,1)$ is assigned independently at random. Further assume that $ \mathbb{E}[\omega_{ij}^2]/\mathbb{E}[\omega_{ij}] \geq 1 $ and $\mathbb{E}[\omega_{ij}^2]/\mathbb{E}^2[\omega_{ij}] < c$. Then, the spectrum of $B$, with high probability, can be described as follows:
	\begin{itemize}
		\item there exist only two real eigenvalues in the spectrum of $B$ with modulus greater or equal to one:
		\begin{align}
		\lambda_1 = c\mathbb{E}[\omega]+ o(c)~, \quad \lambda_{-1} = \frac{\mathbb{E}[\omega^2]}{\mathbb{E}[\omega]} + o(1). \label{eq:eigs}
		\end{align}
		The eigenvalue $\lambda_1$ is the largest in modulus in the spectrum of $B$;
		\item all eigenvalues with non-zero imaginary part have a modulus bounded by $R = \sqrt{c\mathbb{E}[\omega^2]}+ o(\sqrt{c})$.
	\end{itemize}
\label{conj:1}
\end{claim}

Note that Claim~\ref{conj:1} does not make the assumption that $c\to \infty$ as $n \to \infty$, nor that $c = O_n(1)$. Extensive simulations indeed concur in suggesting that the claim holds in both dense and sparse graph regimes. The claim is thus stated for \emph{any} average degree, so long that the underlying graph is of the Erd\H{o}s-R\'enyi type. In detail, the technical condition $\mathbb{E}[\omega_{ij}^2]/\mathbb{E}^2[\omega_{ij}] < c$ is set to enforce that the leading eigenvalue $\lambda_1$ is greater than the radius of the bulk spectrum (hence that it is isolated) and that $\lambda_{-1}$ is smaller than the radius of the bulk: a transition occurs at $\mathbb{E}[\omega_{ij}^2]/\mathbb{E}^2[\omega_{ij}] = c$ where both eigenvalues coincide: $\lambda_1 = \lambda_{-1}$. This inequality condition will thus ensure, when it comes to statistical inference, that non-trivial $\bm\sigma\in\{\pm 1\}^n$ configurations can be theoretically recovered (\emph{i.e.}, that the inference problem is feasible). 
As a practical support to Claim~\ref{conj:1}, Figure~\ref{fig:spec_B} displays the spectrum of the matrix $B$ in both moderately dense ($c \sim {\rm log}^2(n)$) and sparse ($c = O_n(1)$) regimes. 

\bigskip

The fundamental corollary of Claim~\ref{conj:1} is that, in the case of present interest where $\omega_e = {\rm th}(\beta J_e)$, from Equation~\eqref{eq:eigs}, the \emph{inner} eigenvalue $\lambda_{-1}$ of $B$ is equal to 
$$\lambda_{-1} = \frac{\mathbb{E}[{\rm th}^2(\beta J)]}{\mathbb{E}[{\rm th}(\beta J)]} + o(1).$$ 
Exploiting Property~\ref{prop:th}, it follows immediately that, at $\beta = \beta_{\rm N}$, 
$$\lambda_{-1} \underset{\beta=\beta_{\rm N}}{=} 1 + o(1).$$ 
Tuning the value of $\beta$ until $\lambda_{-1} = 1$ thus provides \emph{a method to estimate $\beta_{\rm N}$}. The question on how to efficiently exploit this essential remark from an algorithmic standpoint will be further discussed in Section~\ref{sec:main.algo}. 

\medskip

Before pushing further our main line of deductions, we first introduce some arguments in support of Claim~\ref{conj:1}, which we provide first in the dense and then in the sparse regimes. These are ``arguments'' in the sense that they lack of full mathematical rigour and do not provide a formal proof of Claim~\ref{conj:1}. Specifically, for the dense regime, we adopt a perturbative approach in which we heuristically show that the eigenvalues of $B$ with modulus greater than one are close to the eigenvalues of the (easy to study) matrix $M_0$ appearing in Equation~\eqref{eq:M0}. In the sparse regime, instead, we note that the position of the largest isolated eigenvalue and the radius of the bulk of $B$ obtained in the dense case match the rigorous results of \citep{stephan2020non} proved for the sparse regime. With the support of extensive numerical simulations, we conjecture that the same result obtained in the dense regime to describe the \emph{inner} isolated eigenvalue holds in the sparse regime as well.

\subsubsection{Arguments in support of Claim~\ref{conj:1}}

\textbf{Dense graphs}

\medskip

We first consider a dense graph regime, \emph{i.e.}, when the average degree $c$ goes to infinity faster than ${\rm log}(n)$. This argument is inspired from the proof provided in \citep{coste2019eigenvalues} for unweighted dense graphs with a community structure. The proof of \citep{coste2019eigenvalues} can be straightforwardly adapted to the \emph{binary} case in which $f(e) \in \{\pm \omega\}$, but does not unfold so directly for generic functions $f$.

The main advantage of the dense regime follows from the fact that the degree distribution of $\mathcal{G}$ is almost regular and the Erd\H{o}s-R\'enyi graph is close to a $c$-regular graph \citep{bollobas2001random}, the analysis of which is easier to handle.
This makes it possible to relate the eigenvalues of $B$ to those of $W \in \mathbb{R}^{n \times n}$, defined as $W_{ij} = \omega_{ij}$ if $(ij) \in \mathcal{E}$ and zero otherwise. The idea is to create a sequence of matrices $M(\bm{g}) \in \mathbb{R}^{2n\times 2n}$ (one for each eigenvector $\bm{g}$ of $B$), in the spirit of a proof proposed by Bass of the celebrated Ihara-Bass formula \citep{horton2006zeta}, and to show that all the eigenvalues of $M(\bm{g})$ can be approximated, in the large $n$ limit, by the eigenvalues of a common matrix $M_0$ independent of $\bm{g}$, so long that $\bm{g}$ is an eigenvector corresponding to an eigenvalue $\lambda$ of $B$ for which $|\lambda| \geq 1$. It is the precise study of the spectrum of the limiting $M_0$ which induces the results of Claim~\ref{conj:1}.

\medskip

More specifically, let $\bm{g} \in \mathbb{C}^{2|\mathcal{E}|}$ be an eigenvector of $B$ with eigenvalue $\lambda$, satisfying $|\lambda| \geq 1$ and let $\bm{\omega} \in \mathbb{R}^{2|\mathcal{E}|}$ be the vector containing the weights of the non-zero entries of the matrix $B$ (and recall that $\omega_{ij} = \omega_{ji}$).
Then define the vectors $\bm{\psi}(\bm{g}), \bm{\tilde{\psi}}(\bm{g}) \in \mathbb{C}^n$ as
\begin{align}
\psi_i(\bm{g}) = \sum_{j \in \partial i} \omega_{ij}g_{ij} ~&,~ \quad
\tilde{\psi}_i(\bm{g}) = \sum_{j \in \partial i} \omega^2_{ij}g_{ji} \label{eq:psi}
\end{align}
and $F(\bm{g}) \in \mathbb{C}^{2n \times 2n}$ be any matrix satisfying 
\begin{align}
\big[F(\bm{g})\bm{\psi}(\bm{g})\big]_i &= \sum_{j \in \partial i} \omega_{ij}^3 g_{ij} .
\label{eq:F}
\end{align}
We now wish to relate the quantities $\bm{\psi}(\bm{g}), \bm{\tilde{\psi}}(\bm{g}), F(\bm{g})$ to the eigenvalues of $B$. In particular,
\begin{align*}
\lambda\psi_i(\bm{g}) &= \psi_i(B\bm{g}) = \sum_{j \in \partial i} \omega_{ij} \sum_{(k\ell)} \delta_{jk}(1-\delta_{i\ell})\omega_{kl}g_{k\ell} = \sum_{j \in \partial i} \omega_{ij}\left[\sum_{l \in \partial j} \omega_{j\ell}g_{j\ell} - \omega_{ji}g_{ji}\right] \nonumber \\
&= \left[W\bm{\psi}(\bm{g})\right]_i - \tilde{\psi}_i(\bm{g})
\end{align*}
and, similarly,

\begin{align*}
\lambda\tilde{\psi}_i(\bm{g}) &= \tilde{\psi}_i(B\bm{g}) = \sum_{j \in \partial i} \omega_{ij}^2 \sum_{(k\ell)}\delta_{ik}(1-\delta_{j\ell})\omega_{k\ell}g_{k\ell} = \sum_{j \in \partial i} \omega^2_{ij}\left[\sum_{\ell \in \partial i} \omega_{i\ell}g_{i\ell} - \omega_{ij}g_{ij}\right] \nonumber \\
&= \left[D_W \bm{\psi}(\bm{g})\right]_i - \left[F(\bm{g})\bm{\psi}(\bm{g})\right]_i,
\end{align*}
where $D_W \in \mathbb{R}^{n \times n}$ is the diagonal matrix with $[D_{W}]_{ii} = \sum_{j \in \partial i} \omega^2_{ij}$. Thus, the eigenvalue $\lambda$ is also an eigenvalue of the matrix
\begin{align}
M(\bm{g}) = \begin{pmatrix}
W & -I_n \\
D_W - F(\bm{g}) & 0
\end{pmatrix}.
\label{eq:Mg}
\end{align}

The main difficulty of the analysis is of course introduced by the matrix $F(\bm{g})$ which is different for each eigenvector of $B$ associated to $|\lambda|>1$. In the \emph{binary} case in which $W_{ij} \in \{\pm \omega\}$ for all $(ij) \in\ \mathcal{E}$, this term simplifies: combining Equations~\eqref{eq:psi} and \eqref{eq:F}, we get $F(\bm{g})\bm{\psi} = \omega^2 \bm{\psi}$ and $F(\bm{g})$ thus simplifies \emph{for all $\bm{g}$} into $F(\bm{g}) = \omega^2 I_n$; this allows for a straightforward adaptation of the proof of \cite{coste2019eigenvalues}. The non-binary case is, however, more involved, but the term $\big(D_W - F(\bm{g})\big)\bm{\psi}$ is still dominated by the action of $D_W$:
\begin{align*}
\left|\frac{[F(\bm{g})\bm{\psi}(\bm{g})]_i}{\psi_i(\bm{g})}\right| = \left|\frac{\sum_{j \in \partial i} \omega_{ij}^3 g_{ij}}{\sum_{j \in \partial i} \omega_{ij}g_{ij}}\right|  = o(c).
\end{align*}

For the last equality, we exploited the fact that $\omega_{ij}$ and $\omega_{ij}^3$ are both bounded in $(-1,1)$ and have the same sign: this step is reasonable but non-rigorous, the main theoretical difficulty arising from the dependence between $\omega_{ij}$ and $g_{ij}$. Consequently, $F(\bm{g})$ can be regarded as a small perturbation of $D_W$. Further exploiting the concentration of the degrees, one may thus write
\begin{align*}
\left\Vert \Big(D_W - F(\bm{g})\Big) - c\mathbb{E}[\omega^2]I_n\right\Vert = o(c).
\end{align*}

The eigenvalues of $M(\bm{g})$ can therefore be approximated by those of the matrix
\begin{align}
M_0 = \begin{pmatrix}
W & -I_n \\
c\mathbb{E}[\omega^2]I_n & 0
\end{pmatrix}.
\label{eq:M0}
\end{align}

The spectrum of $M_0$ is trivially related to the spectrum of $W$. Letting $\{\mu_i\}_{i = 1,\dots,n}$ be the eigenvalues of $W$ and $\{\lambda_{0,i}\}_{i = \pm 1, \dots, \pm n}$ those of $M_0$, by the block determinant formula \citep[Section 5]{silvester2000determinants}, it comes that
\begin{align}
\lambda_{0,\pm i} = \frac{\mu_i \pm \sqrt{\mu_i^2-4c\mathbb{E}[\omega^2]}}{2}.
\label{eq:M0_eigs}
\end{align}

\begin{figure}
	\centering
	\includegraphics[width=0.7\columnwidth]{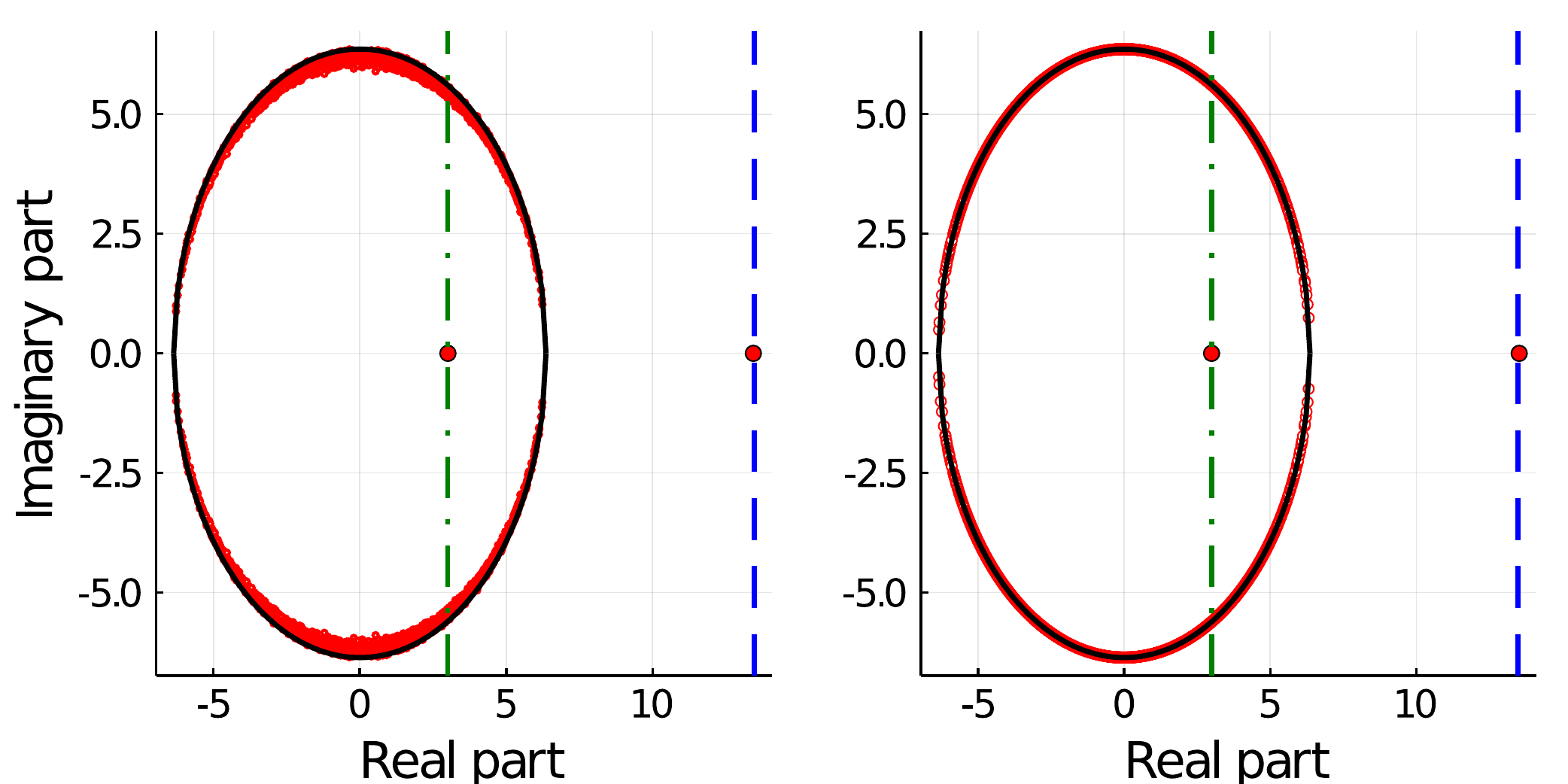}
	\caption{\textbf{Left}: spectrum of the matrices $M(\bm{g})$ defined in Equation~\eqref{eq:Mg} with $\bm{g}$, one of the eigenvectors of $B$ attached to a complex eigenvalue. \textbf{Right}: Spectrum of $M_0$, defined in Equation~\eqref{eq:M0}. The graph considered is the same for the two matrices, with $n = 1~500$, $c = {\rm log}^2(n)$. The matrix $W = {\rm th}(\beta J)$, with $\beta = 1$ and the entries $J_{ij}$ are i.i.d. normal variable with $J_0 = 1$ and $\nu = 3$. The blue dotted line is the position of $c\mathbb{E}[{\rm th}(\beta J)]$, the green dash-dotted line is the position of $\mathbb{E}[{\rm th}^2(\beta J)]/\mathbb{E}[{\rm th}(\beta J)]$, while the black solid line is the circle in the complex plane of radius $\sqrt{c\mathbb{E}[{\rm th}^2(\beta J)]}$.}
	\label{fig:MMg}
\end{figure}

In particular, it unfolds that
\begin{align*}
\mu_i^2 \geq 4c\mathbb{E}[\omega^2] \quad &\Longrightarrow \quad \lambda_{0,-i} = \frac{c\mathbb{E}[\omega^2]}{\lambda_{0,i}} \equiv \frac{R^2}{\lambda_{0,i}} \\
\mu_i^2 < 4c\mathbb{E}[\omega^2] \quad &\Longrightarrow \quad |\lambda_{0,\pm i}| = \sqrt{c\mathbb{E}[\omega^2]} \equiv R .
\end{align*}

Applying successively Wigner's semi-circle theorem \citep{wigner1958distribution} and Bauer-Fike's theorem \citep{bauer1960norms}, we thus have that
\begin{align}
\mu_1 = c\mathbb{E}[\omega] + o(\sqrt{c}), \quad |\mu_i|_{i \geq 2} \leq \sqrt{c\mathbb{E}[\omega^2]} + o(\sqrt{c}).
\label{eq:wigner}
\end{align}

Combining Equation~\eqref{eq:M0_eigs} and \eqref{eq:wigner}, along with the fact that the eigenvalues $\{\lambda_{0,\pm i}\}_{i = 1,\dots, n}$ are a close approximation to the eigenvalues of $B$ with modulus greater than one, we obtain the formulation of Claim~\ref{conj:1}. Figure~\ref{fig:MMg} compares the spectra of the matrices $M(\bm{g})$ and $M_0$, which should be themselves compared to the left display in Figure~\ref{fig:spec_B}. Appendix~\ref{app:F_matrix} provides the explicit expressions of the matrix $F(\bm{g})$ used in Figure~\ref{fig:MMg}.   

\medskip

This technical argument provides important intuitions on the spectrum of $B$: i) the leading eigenvalue of $B$ (the largest in modulus) is determined by the expectation of the entries of $W$; ii) the radius of the bulk of $B$ is determined by the expectation of the squared entries of $W$; iii) all eigenvalues of $B$ with modulus greater than one come in pairs (see Equation~\eqref{eq:M0_eigs}): they are complex conjugates if their imaginary part is non-zero or \emph{harmonic conjugate} if they are real. 
This last observation justifies the existence of a real isolated eigenvalue \emph{inside} the bulk of $B$, the importance of which will be further discussed in Section~\ref{sec:main.phd}.

\medskip

As a downside, the setting considered in this section is, somehow, too simplistic. The analysis allowed us to neglect the term $F(\bm{g})$, which plays the role of the \emph{Onsager reaction term} \citep{mezard1987spin} which does not appear in the na\"ive mean-field approximation but plays a crucial role in the Bethe approximation. The fact that $F(\bm{g})$ can be neglected thus indicates that the regime under consideration is somehow \emph{too simple}, the spectral behavior of $B$ being fully determined by $W$.

\medskip

Consequently, we next discuss the far more interesting \emph{sparse} regime in which the \emph{Onsager reaction term} plays a fundamental role and in which the Bethe approximation brings a decisive advantage over the na\"ive mean-field approximation. In the sparse regime, the structure of the spectrum of the matrix $B$ is essentially preserved, as well as the fact that all its eigenvalues all come in real harmonic or complex conjugate pairs.

\bigskip

\textbf{Sparse graphs}

\medskip

The Bethe approximation is exact on trees \citep{mezard2009information} and asymptotically yields (in the large $n$ limit) exact results on locally \emph{tree-like} graphs. This is precisely the case of sparse Erd\H{o}s-R\'enyi graphs, in which the average degree is of order $c = O_n(1)$. In this case, the spectrum of the matrix $W$ is no longer formed by an isolated eigenvalue (the largest in modulus) and a \emph{bulk} of eigenvalues close to each other that follow the semi-circle law, as it happens in the \emph{dense} regime discussed in the previous paragraph. Here the eigenvalues of $W$ are known to have an unbounded support (little else is in fact theoretically known about this spectrum).

\medskip

The non-backtracking matrix $B$, instead, essentially preserves the same spectral structure as in the dense regime, in which the \emph{bulk} eigenvalues are bounded by a circle in the complex plane, as shown in Figure~\ref{fig:spec_B}. This result was recently proved in \citep{stephan2020non} in which the authors showed that, under the assumptions of Claim~\ref{conj:1}, the matrix $B$ has an isolated eigenvalue equal to $\lambda_1 = c\mathbb{E}[\omega] + o_n(1)$, (recall that $c = O_n(1)$) while all other eigenvalues satisfy $|\lambda_{i\geq 2}| \leq \sqrt{c\mathbb{E}[\omega^2]} + o(1)$. The result of \citep{stephan2020non} however does not mention the existence of \emph{inner} real eigenvalues in the spectrum of $B$ and, to best of our knowledge, no mathematical tool has been developed yet to rigorously address this question in the sparse regime. Yet, the position of the leading eigenvalue of $B$ and the radius of its bulk spectrum are the same as in the \emph{dense} graph case. We then conjecture, supported by extensive simulations, that also the inner isolated eigenvalue has the same position as in the dense regime, given by the square radius of the bulk, divided by the leading eigenvalue of $B$.

\medskip

We take the opportunity of the reference to \citep{stephan2020non} to generalize the central claim of the article to their richer context. This result is of independent interest, particularly for more structured graph models.

\begin{rem}[Random sparse graphs with independent entries]
	Note that the result of \citep{stephan2020non} is given under more general hypotheses than those discussed here. Specifically, the authors of \citep{stephan2020non} consider a setting in which each edge of the graph $\mathcal{G}$ is created independently at random with probability $p_{ij}$. The Erd\H{o}s-R\'enyi graph falls into the particular case in which $P = \{p_{ij}\}_{i,j=1}^n = \frac{c}{n}\bm{1}_n\bm{1}_n^T$. The leading (real) eigenvalues of $B$ are determined from the leading eigenvalues of $P\circ \mathbb{E}[W]$, and the bulk radius by the leading eigenvalue of $P\circ \mathbb{E}[W\circ W]$. Based on this result, we conjecture that the real eigenvalues of $B$ come in harmonic pairs precisely determined by
	\begin{align*}
	\lambda_{\pm i} = \frac{\rho\left(P\circ \mathbb{E}[W \circ W]\right)}{\gamma_i\left(P \circ\mathbb{E}[W]\right)},
	\end{align*}
	where $\rho(\cdot)$ indicates the largest eigenvalue in modulus, and $\gamma_i(P \circ \mathbb{E}[W])$ are the eigenvalues of $P\circ \mathbb{E}[W]$, greater than $\sqrt{\rho\left(P\circ \mathbb{E}[W \circ W]\right)}$. 
	
	A particular case of this setting is the \emph{degree-corrected stochastic block model} which reproduces a $k$-class structure on an unweighted graph. In this case, the matrix $\mathbb{E}[W]$ has a low rank factorization $\mathbb{E}[W] = \frac{1}{n}\sum_{i = 1}^k \alpha_i \bm{u}_i\bm{u}_i^T$, with $\alpha_1 = c$. Furthermore $P = \bm{\theta}\bm{\theta}^T$, where $\bm{\theta}^T\bm{1}_n = n$ and $\frac{1}{n}\bm{\theta}^T\bm{\theta} := \Phi$. Then, in agreement with \citep{gulikers2016non} and the conjecture of \citep{dall2019revisiting}, the eigenvalues of $B$ can be described as follows
	\begin{align*}
	\lambda_{i} &= \alpha_i\Phi + o_n(1)\quad {\rm for}~ 1\leq i\leq k \\
	\lambda_{-i} &= \frac{c}{\alpha_i} + o_n(1)\quad {\rm for}~1\leq i \leq k \\
	|\lambda_{i > k}| &\leq \sqrt{c\Phi} + o_n(1).
	\end{align*}
\end{rem}

\medskip

Returning to the implications of Claim~\ref{conj:1} of immediate interest, recall that the claim makes it possible to relate the Nishimori temperature to the specific eigenvalue $\lambda_{-1}$ of the matrix $B$. From a numerical standpoint though, $\lambda_{-1}$ is not easily accessible for two reasons: i) the matrix $B$ is non-symmetric and large, slowing down eigenvalue computations; ii) since $\lambda_{-1}$ is smaller in modulus than most of the complex eigenvalues of $B$, while not being the smallest in modulus (see Figure~\ref{fig:spec_B}), one needs to compute all the bulk eigenvalues of $B$ in order to access $\lambda_{-1}$: this comes at an impractical computational cost of $O(cn^3)$ with state of the art methods \citep[see for example]{saad1992numerical}.
We next show that, as a consequence of Claim~\ref{conj:1} and Property~\ref{prop:th}, the (symmetric) Bethe-Hessian matrix $H(x)$ \eqref{eq:BH} can be efficiently used to estimate $\beta_{\rm N}$ in the RBIM with a computational cost scaling as $O(nc)$.

\subsection{The relation between $\beta_{\rm N}$ and the Bethe-Hessian matrix}
\label{sec:main.phd}

This section elaborates on our final relation between the Bethe-Hessian matrix and the Nishimori temperature, as well as on how the respective spectra of the matrices $H(x)$ and $B$ can be related to the phase diagram of Figure~\ref{fig:ph_d}.

\subsubsection{The Bethe free energy}
\label{sec:variational}

Let us first recall the basics of a variational approach, and specifically of the Bethe approximation. For $\mu(\bm{s})$ the Boltzmann distribution \eqref{eq:Boltzmann}, the free energy $F_{J,\beta}$ and the variational free energy $\tilde{F}_{J,\beta}(\bm{q})$ (given for an arbitrary set of parameters $\bm{q}$), are defined through
\begin{align}
F_{J,\beta} &= \sum_{\bm{s}} \mu(\bm{s})\Big(\beta\mathcal{H}_J(\bm{s}) + {\rm log}~\mu(\bm{s})\Big) \label{eq:free_en}\\
\tilde{F}_{J,\beta}(\bm{q}) &= \sum_{\bm{s}} p_{\bm{q}}(\bm{s})\Big(\beta\mathcal{H}_J(\bm{s}) + {\rm log}~p_{\bm q}(\bm{s})\Big). \label{eq:free_var}
\end{align}

The function $F_{J,\beta}$ is a moment generating function for the 
Boltzmann distribution of Equation~\eqref{eq:Boltzmann_distr} but, in general, cannot be computed exactly. The variational free energy $\tilde{F}_{J,\beta}(\bm{q})$ represents a tractable approximation of $F_{J,\beta}$.
From a straightforward calculation it can in particular be shown that $\tilde{F}_{J,\beta}(\bm{q}) - F_{J,\beta} = D_{\rm KL}(\mu || p_{\bm q}) \geq 0$, where $D_{\rm KL}(\cdot || \cdot)$ is the Kullback-Leibler divergence between two distributions. For a parametrized family of distributions $p_{\bm q}$, minimizing the variational free energy with respect to $\bm{q}$ provides the Kullbach-Liebler optimal approximation of $F_{J,\beta}$. The variational Bethe approximation considers a mean- and covariance-parametrized distribution $p_{\bm q}=p_{\bm{m},\bm{\chi}}$ defined as
\begin{align}
p_{\bm{m},\bm{\chi}}(\bm{s}) = {\prod_{(ij)\in\mathcal{E}} \frac{1+m_is_i+m_js_j+\chi_{ij}s_is_j}{4}}\cdot \prod_{i = 1}^n \left[\frac{1+m_is_i}{2}\right]^{1-d_i},
\label{eq:B_app}
\end{align}
where $m_i$ and $\chi_{ij}$ are the average of $s_i$ and $s_is_j$ according to $p_{\bm{m},\bm{\chi}}(\bm{s})$, respectively. Here $d_i$ denotes the degree of node $i$ ($d_i~=~|\{j~:~(ij) \in \mathcal{E}\}|$). The approximation turns out to be the exact factorization of $\mu(\bm{s})$ when $\mathcal{G}$ is a tree, and is thus often claimed a good approximation of it in sparse, tree-like graphs.

\medskip

A complete derivation of the Bethe-Hessian matrix from the Bethe free energy is proposed in \citep{watanabe2009graph}. It is instructive though to recall its main steps which allow one to relate the Bethe-Hessian matrix eigenvalues to the phase diagram of Figure~\ref{fig:ph_d}.
From the expression of $\tilde{F}^{\rm Bethe}_{J,\beta}(\bm{m},\bm{\chi})$, obtained combining Equations~(\ref{eq:free_var}, \ref{eq:B_app}), one obtains that $\nabla_{\bm{m}} \tilde{F}^{\rm Bethe}_{J,\beta}(\bm{m},\bm{\chi})|_{\bm{m} = 0} = 0$, \emph{i.e.}, the \emph{paramagnetic} point is always an extremum of the Bethe free energy. In order to study the stability of this solution, we consider the \emph{Hessian} matrix of the variational free energy, computed at the paramagnetic point: the smallest eigenvalues of this matrix are associated to the \emph{local} directions along which the paramagnetic solution may get unstable and non-trivial order in the spin configurations can be observed. This Hessian matrix explicitly reads:
\begin{align}
\left. \frac{\partial^2 \tilde{F}^{\rm Bethe}_{J,\beta}(\bm{m},\bm{\chi})}{\partial m_i \partial m_j} \right|_{\bm{m} = 0} = \delta_{ij}\left(1+\sum_{k \in \partial i} \frac{\chi_{ik}^2}{1-\chi^2_{ik}}\right) - \frac{\chi_{ij}}{1-\chi_{ij}^2}.
\label{eq:real_BH}
\end{align} 

By further computing the gradient of $\tilde{F}^{\rm Bethe}_{J,\beta}(\bm{m},\bm{\chi})$ with respect to $\bm{\chi}$, one next obtains $\chi_{ij} = {\rm th}(\beta J_{ij})$ (as already mentioned in the Section~\ref{sec:basic} where we pointed that, in the paramagnetic phase, $\langle s_is_j \rangle_{\beta} = {\rm th}(\beta J_{ij})$ under the Bethe approximation).
Setting $\omega_{ij} = {\rm th}(\beta J_{ij})$, the matrix of Equation~\eqref{eq:real_BH} precisely corresponds to $H(1)$ defined in Equation~\eqref{eq:BH}. We denote this matrix $H_{\beta, J}$, which explicitly reads:
\begin{align}
\left(H_{\beta, J}\right)_{ij} =\delta_{ij}\left(1+\sum_{k \in \partial i}\frac{{\rm th}^2(\beta J_{ik})}{1-{\rm th}^2(\beta J_{ik})}\right) - \frac{{\rm th}(\beta J_{ij})}{1-{\rm th}^2(\beta J_{ij})}.
\label{eq:BH_th}
\end{align}

\medskip

We may now relate the Bethe approximation to the phase diagram of Figure~\ref{fig:ph_d}.

\begin{figure}[t!]
	\centering
	\includegraphics[width=\columnwidth]{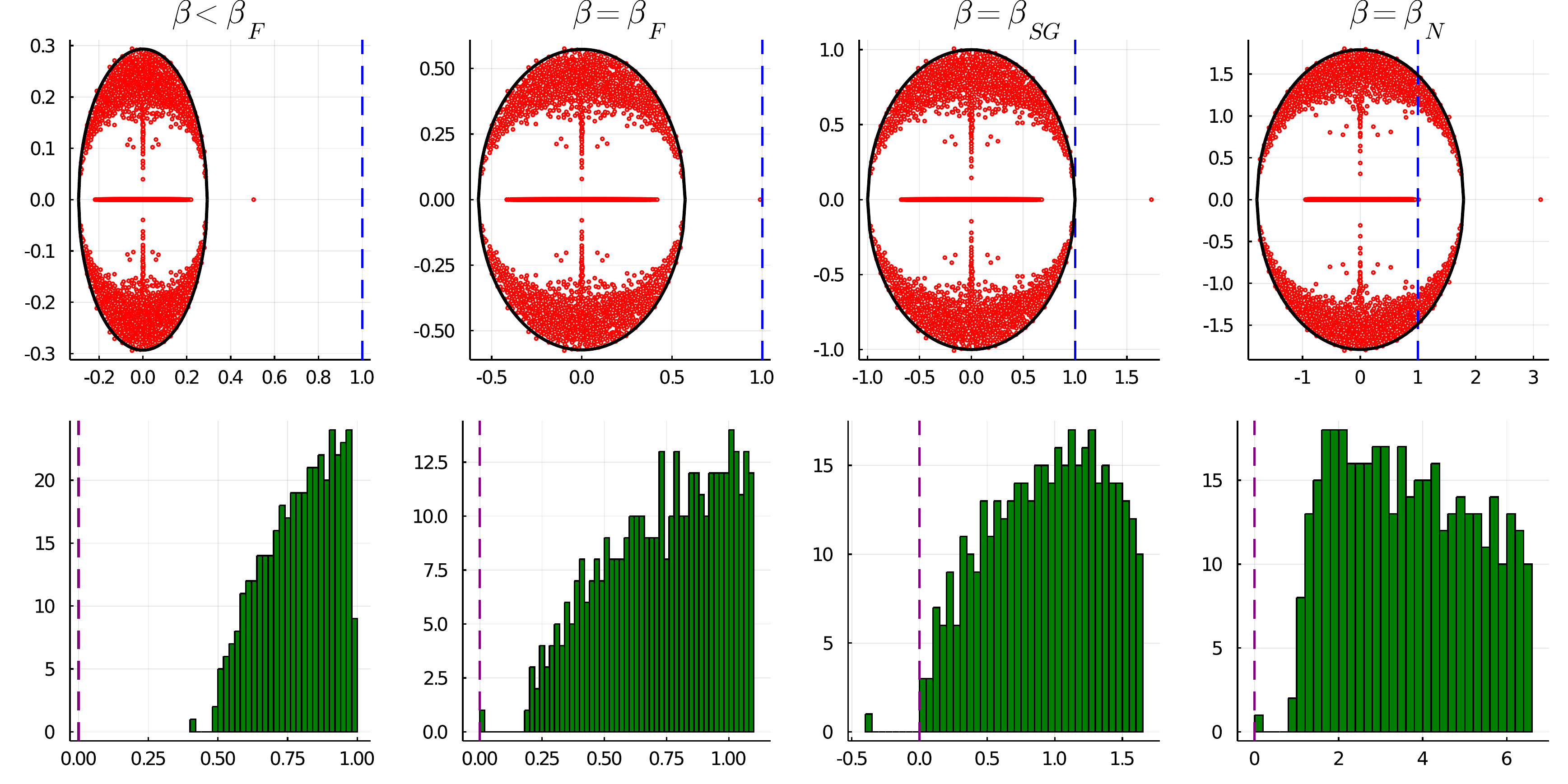}
	\caption{\textbf{First row}: spectrum of the matrix $B$ in the complex plane for different values of $\beta$; \textbf{Second row}: histogram of the eigenvalues of $H_{\beta,J}$ (zoomed in on the smallest eigenvalues) for different values of $\beta$. 
		\textbf{First column}: $\beta = 0.5 \beta_{\rm F}$, paramagnetic phase; \textbf{Second column}: $\beta = \beta_{\rm F}$ paramagnetic-ferromagnetic transition; \textbf{Third column}: $\beta = \beta_{\rm SG}$ paramagnetic- spin glass phase transition; \textbf{Fourth column}: $\beta = \beta_{\rm N}$, Nishimori temperature. For all matrices, the same graph was used with $n = 1\,000$, $c = 10$. The weights of the edges are $\omega_{ij} = {\rm th}(\beta J_{ij})$ for the different values of $\beta$ just described. The $J_{ij}$ are drawn independently from a Gaussian distribution with $J_0 = 1$ and $\nu = 1.5$. The blue lines in the first row is the vertical line at $x=1$, while the purple line in the second row is the vertical line at $x = 0$.}
	\label{fig:spec_phd}
\end{figure}

\subsubsection{Phase diagram}

Let us move back to the system described by Equations~(\ref{eq:ham}, \ref{eq:Boltzmann}) and Definition~\ref{def:Nish}, first set at sufficiently high temperature (small $\beta$). In this case, for all $\beta_{\rm N}$ the system is in the \emph{paramagnetic} phase, for which $\langle s_i \rangle_{\beta} = 0$. The paramagnetic solution $\bm{m} = 0$ is a minimum of $\tilde{F}_{J,\beta}^{\rm  Bethe}(\bm{m},\bm{\chi})$,  $H_{\beta,J}$ is positive definite.

Consider now $\beta_{\rm N}$ to be sufficiently large, so that the system undergoes to a \emph{paramagnetic-ferromagnetic} phase transition (see Figure~\ref{fig:ph_d}). For $\beta = \beta_{\rm F}$ defined as $c\mathbb{E}[{\rm th}(\beta_{\rm F} J)] = 1$, the leading eigenvalue of $B$ is equal to $1$ and one of the eigenvalues of $H_{\beta, J}$ is equal to zero. This eigenvalue is necessarily the smallest, since for $\beta < \beta_{\rm F}$ all the eigenvalues are positive.

For small values of $\beta_{\rm N}$, the system undergoes the \emph{paramagnetic-spin glass} phase transition (see Figure~\ref{fig:ph_d}) at the temperature $\beta = \beta_{\rm SG}$ defined so that $c\mathbb{E}[{\rm th}^2(\beta_{\rm SG} J)] = 1$. For this value of $\beta$, the radius of the bulk of the matrix $B$ is equal to one and the bulk of $H_{\beta,J}$ is asymptotically close to zero.

Finally, further decreasing the temperature, at $\beta = \beta_{\rm N}$ defined by $\mathbb{E}[{\rm th}^2(\beta_{\rm N}J)] = \mathbb{E}[{\rm th}(\beta_{\rm N}J)]$, the eigenvalue $\lambda_{-1}$ is equal to one and the smallest eigenvalue of $H_{\beta,J}$ reaches zero for the second time. 
In Figure~\ref{fig:spec_phd} we show the spectra of the matrices $B$ and $H_{\beta,J}$ at $\beta < \beta_{\rm F}$, $\beta = \beta_{\rm F}, \beta_{\rm SG}, \beta_{\rm N}$ that  confirm the relation between the spectra of these two matrices and the phase diagram.

\medskip

Having established in depth the relation between the matrices $B$ and $H(x)$, and their relations to the phase diagram, we now show how one can efficiently estimate $\beta_{\rm N}$, exploiting the smallest eigenvalue of $H_{\beta, J}$.

\subsection{Estimation of $\beta_{\rm N}$ from $H_{\beta,J}$}
\label{sec:main.algo}

The present section provides a numerically efficient estimator $\hat{\beta}_{\rm N}$ of the Nishimori temperature, first defined formally and then under the form of the output of a practical \emph{algorithm}.

The proposed value of $\hat{\beta}_{\rm N}$, estimate of the genuine Nishimori temperature $\beta_{\rm N}$, reads
\begin{align}
\hat{\beta}_{\rm N} = \underset{\beta}{\rm max}\left\{\beta~:~\gamma_{\rm min}(H_{\beta, J}) = 0\right\},
\label{eq:beta*}
\end{align}   
where $\gamma_{\rm min}(\cdot)$ indicates the smallest eigenvalue of a matrix.
Under this definition, not only does $\hat{\beta}_{\rm N}$ provides a consistent estimate of $\beta_{\rm N}$ for $J$ distributed as Definition~\ref{def:Nish}, this being a consequence of Claim~\ref{conj:1}, but it also provides the ``best guess'' of an hypothetically corresponding $\beta_{\rm N}$ for matrices $J$ which would follow a different distribution from the model of Definition~\ref{def:Nish}. Indeed, $\hat{\beta}_{\rm N}$ has the advantage of always being defined, even for arbitrary matrices $J$, while having a clear interpretation for the class of matrices that fall under Definition~\ref{def:Nish}. This definition is particularly reminiscent of the algorithm proposed in \citep{dall2020unified} for community detection over sparse heterogeneous graphs, and which demonstrates a robust behavior on applications to real-world graphs.

\begin{figure}[t!]
	\centering
	\includegraphics[width=\columnwidth]{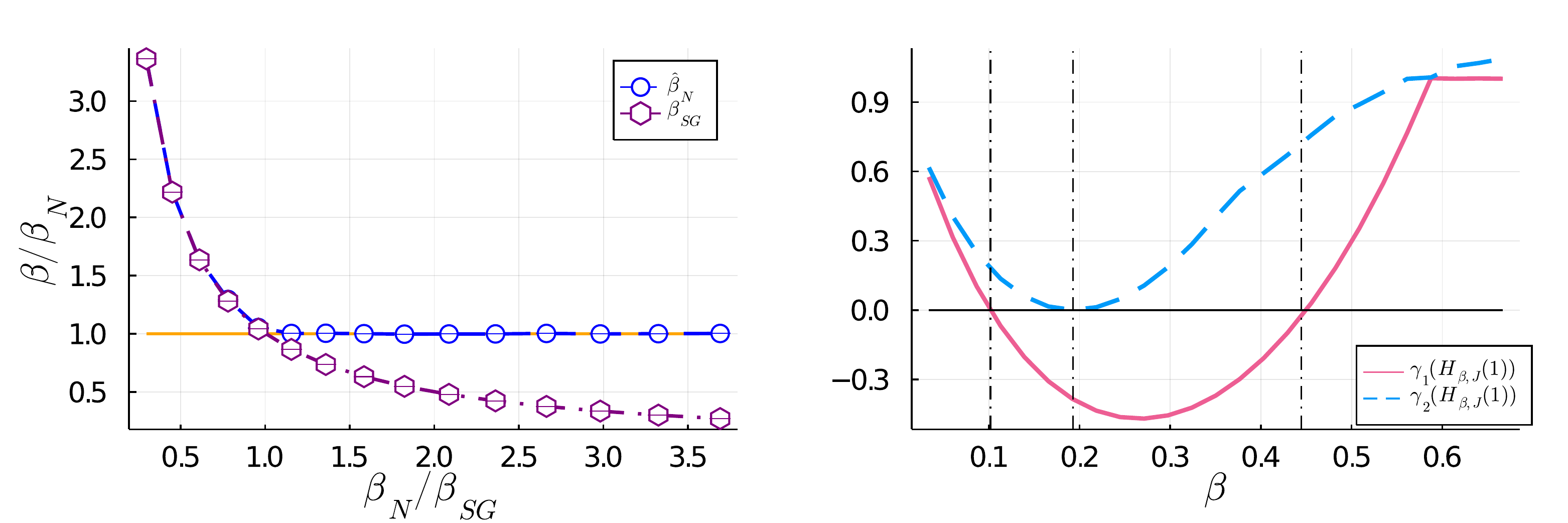}	
	\caption{\textbf{Left}: computation of $\hat{\beta}_{\rm N}$ for different values of $\beta_{\rm N}$. The blue dots represent the ratio between $\hat{\beta}_{\rm N}$, computed with Algorithm \ref{alg:1} and the analytical value of $\beta_{\rm N}$. The purple hexagons are the value of $\beta_{\rm SG}/\beta_{\rm N}$, while the orange line is at $y = 1$. For these plots, $n = 10\,000$ and $c = 5$. The weights of the non-zero entries of $J$ are distributed i.i.d.\@ according to $\mathcal N(J_0,\nu^2)$ for $J_0$ ranging from $J_0 = 0.5$ to $J_0 = 4$ and $\nu = 3.5$. Averages are taken over ten samples. \textbf{Right}: behavior of the two smallest eigenvalues of $H_{\beta,J}$ as a function of $\beta$. The solid line indicates the smallest eigenvalue, while the dotted line is the second smallest. The vertical lines are set at $\beta_{\rm F} < \beta_{\rm SG} < \beta_{\rm N}$. For this simulation, $n = 30\,000$ and $c=10$. The weights of the matrix $J$ are distributed i.i.d.\@ according to a $\mathcal N(J_0,\nu^2)$ with $J_0=1$ and $\nu = 1.5$.}
	\label{fig:eig_beta}
\end{figure}

\medskip

To best understand the rationale behind the definition of $\hat{\beta}_{\rm N}$, first observe that $H_{\beta, J}$ is positive definite for all small values of $\beta$ ($\lim_{\beta \to 0} H_{\beta,J} = I_n$). By increasing $\beta$, the smallest eigenvalue eventually hits zero a first time before turning negative: the zero-crossing occurs precisely at $\beta = \beta_{\rm F}$. Then, continuing increasing $\beta$, at $\beta=\beta_{\rm SG}$, the second smallest eigenvalue of $H_{\beta, J}$ is asymptotically equal to zero and $\gamma_{\rm min}(H_{\beta,J})$ is now negative.
Finally, for $\beta \to \infty$, $H_{\beta, J}$ is again positive definite (the result can be easily obtained using Gershgorin's circle theorem). Therefore, there must exist a value $\beta > \beta_{\rm SG}$ for which $\gamma_{\rm min}(H_{\beta, J}) = 0$ for a second time. This second zero-crossing occurs precisely when $\beta=\hat{\beta}_{\rm N}$. The right display of Figure~\ref{fig:eig_beta} visually explains this behavior.

\medskip

The basic idea of the proposed algorithm to compute $\hat{\beta}_{\rm N}$ precisely consists in starting from $\beta=\beta_{\rm SG}$ to then find the value of $\beta > \beta_{\rm SG}$ for which $\gamma_{\rm min}(H_{\beta,J}) = 0 $.
Following this argument, we propose an iterative algorithm based on Courant-Fischer theorem to compute $\hat{\beta}_{\rm N}$. The output of  Algorithm~\ref{alg:1} is depicted in the left display of Figure~\ref{fig:eig_beta}. Note in particular that, as long as $\beta_{\rm SG} < \beta_{\rm N}$, \emph{i.e.}, so long that $\mathbb{E}[{\rm th}^2(\beta J_{ij})]/\mathbb{E}^2[{\rm th}(\beta J_{ij})] < c$, the value of $\hat{\beta}_{\rm N}$ is a good estimate of $\beta_{\rm N}$. When the condition is instead not met, $\hat{\beta}_{\rm N}$ simply coincides with $\beta_{\rm SG}$. A more detailed analysis of Algorithm~\ref{alg:1} is provided in Appendix~\ref{app:algo}. The numerical advantage of exploiting the Bethe-Hessian matrix is decisive. First $H_{\beta,J}$ is \emph{symmetric} and of size $n\times n$ \emph{regardless of the average node degree}. Most importantly, the only eigenvalue of $H_{\beta,J}$ that needs be computed is the one of smallest amplitude, so that $\hat{\beta}_{\rm N}$ can be estimated at a $O(nc)$ computational cost (using the Arnoldi method \citep{saad1992numerical}).

\medskip

While from a purely physics standpoint,
Claim~\ref{conj:1} is an elegant theoretical relation between the Nishimori temperature and the Bethe-Hessian matrix, when it comes to machine learning applications, estimating $\beta_{\rm N}$ may have practical impact on algorithm performance. In particular, $\hat{\beta}_{\rm N}$ may be used as an approximation of $\beta_{\rm N}$ when solving statistical inference on $\bm\sigma$ (for instance, via an optimal linearization of the Bayes optimal solution) in the absence of knowledge of the parameters in the generative model \eqref{eq:Bayes_opt}.

\begin{algorithm}[!t]
	\SetKwData{Left}{left}\SetKwData{This}{this}\SetKwData{Up}{up}\SetKwFunction{Union}{Union}\SetKwFunction{FindCompress}{FindCompress}\SetKwInOut{Input}{input}\SetKwInOut{Output}{output}
	\Input{Weighted adjacency matrix of a graph $J \in \mathbb{R}^{n \times n}$, precision error $\epsilon \in \mathbb{R}$\;}
	\Output{Value of $\hat{\beta}_{\rm N} \in \mathbb{R}^+$\;}
	\BlankLine
	Compute $c$, the average degree of the underlying unweighted graph: $c=\frac{1}{n}\sum_{i} \sum_{j}\mathbb{I}(J_{ij}\neq 0)$ \;
	Compute $\hat{\beta}_{\rm SG}$ by solving $c\mathbb{E}[{\rm th}^2(\hat{\beta}_{\rm SG}J_{ij})] = 1$\;
	Set $t=1$ and $\beta_{t} \leftarrow \hat{\beta}_{\rm SG}$ \;
	Initialize $\delta\leftarrow+\infty$ \;
	\BlankLine
	\While{$\delta > \epsilon$}{
		Compute $H_{\beta_t, J}$ (Equation~\eqref{eq:BH_th}) \;
		Compute $\gamma_{{\rm min}, t}$, the smallest eigenvalue of $H_{\beta_t, J}$, as well as its associated eigenvector $\bm{x}_t$  \;
		Define the function $f_{t}(\beta') = \bm{x}_t^TH_{\beta',J}\bm{x}_{t}$, for $\beta' \in \mathbb{R}^+$ \;
		Compute $\beta_{t+1}$ by solving $f_{t}(\beta_{t+1}) = 0$ \;
		Update $\delta \leftarrow |\gamma_{{\rm min}, t}|$ \;
		Increment $t \leftarrow t+1$ \;
	}	
	\textbf{return:} $\beta_{t-1}$
	\caption{\texttt{Compute\_}$\hat{\beta}_{\rm N}$}
	\label{alg:1}
\end{algorithm}

\section{Application to node classification}
\label{sec:ml}

This section discusses one of the immediate applications of the results introduced in the previous sections to the context of Bayesian statistical inference, and specifically to the problem of unsupervised node clustering on a graph. To this end, we first establish the relation between the Bayesian optimal inference and the Nishimori temperature, specific to the node classification problem; this then allows us to particularize Algorithm~\ref{alg:1} to this setting.
Possibly most importantly, we conclude by commenting on how the considered model may be extrapolated to perform clustering on (possibly sparse) adjacency matrices of \emph{real data} and relate our resulting proposed algorithm to commonly used competing spectral algorithms.

\subsection{A generative model for node classification}

Let $\mathcal{G}$ be the realization of an Erd\H{o}s-R\'enyi graph whose nodes are divided in two non-overlapping classes, labelled via the vector $\bm{\sigma} \in \{-1,1\}^n$. Associated to $\mathcal{G}$ is a weighted adjacency matrix $\tilde{J} \in \mathbb{R}^{n \times n}$ with probability distribution:
\begin{align}
\mathbb{P}(\tilde{J}|\bm{\sigma}) = \prod_{(ij)\in\mathcal{E}} p_0(|\tilde{J}_{ij}|)e^{\beta_{\rm N}\tilde{J}_{ij}\sigma_i\sigma_j},
\label{eq:Nish_distr_inf}
\end{align}
for an arbitrary non negative function $p_0(\cdot)$ and for some $\beta_{\rm N} > 0$. According to this model, the edges connecting nodes in the same community are more likely to be positive, while those connecting nodes in opposite communities are instead more likely to be negative. Given a realization of $\tilde{J}$, the task of the experimenter (who only has access to $\tilde J$) is to infer the vector $\bm{\sigma}$. We can formulate this problem in terms of a Bayesian inference:
\begin{align}
\mathbb{P}(\bm{\sigma}|	\tilde{J}) = \frac{\mathbb{P}(\tilde{J}|\bm{\sigma})\mathbb{P}(\bm{\sigma})}{\mathbb{P}(\tilde{J})} =  \frac{1}{Z_{\tilde{J}}}~{\rm exp}\left\{\sum_{(ij) \in \mathcal{E}} \beta_{\rm N}\tilde{J}_{ij}\sigma_i\sigma_j\right\}.
\label{eq:inference_Nishi}
\end{align}
Computing the marginals of $\mathbb{P}(\bm{\sigma}|	\tilde{J})$ is equivalent to computing the average magnetization of an Ising model on $\tilde{J}$ at the Nishimori temperature. However, the value of $\beta_{\rm N}$ cannot be easily inferred from $\tilde{J}$ without knowing $\bm{\sigma}$: one would indeed need to solve
\begin{align*}
\mathbb{E}[{\rm th}(\beta_{\rm N} \tilde{J}_{ij}\sigma_i\sigma_j)] = \mathbb{E}[{\rm th}^2(\beta_{\rm N} \tilde{J}_{ij}\sigma_i\sigma_j)].
\end{align*}

To progress further, let us next introduce the matrix $J = \tilde{J} \circ \bm{\sigma}\bm{\sigma}^T$.  Given the probability distribution of $\tilde{J}$ \eqref{eq:Nish_distr_inf}, the matrix $J$ is exactly defined as per Definition~\ref{def:Nish}. The key result to proceed consists in observing that the matrices $H_{\beta,J}$ and $H_{\beta, \tilde{J}}$ have the same eigenvalues and, up to a \emph{gauge} transformation, the same eigenvectors. This then enables the use of Algorithm~\ref{alg:1} to estimate $\beta_{\rm N}$ directly from $\tilde{J}$. Let indeed $\bm{x} \in \mathbb{R}^n$ be an eigenvector of $H_{\beta,\tilde{J}}$ with eigenvalue $\lambda$ and let $\bm{y}$ have entries $y_i = x_i\sigma_i$.  Then
\begin{align*}
\lambda y_i = \lambda x_i\sigma_i 
= \sigma_i\sum_j \left(H_{\beta,\tilde{J}}\right)_{ij}x_j 
= \sigma_i \sum_j \left(H_{\beta,J}\right)_{ij}\sigma_i\sigma_jx_j 
= \left(H_{\beta,J}\bm{y}\right)_i
\end{align*}
so that $\lambda$ is an eigenvalue of $H_{\beta,J}$ with eigenvector $\bm{y}$. Consequently, the smallest eigenvalue of $H_{\beta_{\rm N},\tilde{J}}$ is asymptotically close to zero and Algorithm~\ref{alg:1} can be used to estimate $\beta_{\rm N}$. 

\begin{algorithm}[!t]
	\SetKwData{Left}{left}\SetKwData{This}{this}\SetKwData{Up}{up}\SetKwFunction{Union}{Union}\SetKwFunction{FindCompress}{FindCompress}\SetKwInOut{Input}{input}\SetKwInOut{Output}{output}
	\Input{Weighted adjacency matrix of a graph $\tilde{J} \in \mathbb{R}^{n \times n}$, precision error $\epsilon \in \mathbb{R}$\;}
	\Output{Value of $\hat{\beta}_{\rm N} \in \mathbb{R}^+$, estimated label vector $\bm{\hat{\sigma}} \in \{-1,1\}^n$\;}
	\BlankLine
	Shift the non-zero $\tilde{J}_{ij}$ as: $\tilde{J}_{ij} \leftarrow \tilde{J}_{ij} - \frac{1}{2|\mathcal{E}|}\bm{1}_n^T\tilde{J}\bm{1}_n$ \;
	Compute $\hat{\beta}_{\rm N} \leftarrow$ \texttt{Compute\_}$\hat{\beta}_{\rm N}$ (Algorithm~\ref{alg:1}) \;
	Compute $H_{\hat{\beta}_{\rm N},\tilde{J}}$ (Equation~\eqref{eq:BH_th}) \;
	Compute $\bm{x} \leftarrow$ the eigenvector associated to $\gamma_{\rm min}(H_{\hat{\beta}_{\rm N}, \tilde{J}})$ \;
	Estimate $\bm{\hat{\sigma}}$ as the output of $2$-class \emph{k-means} on the entries of $\bm{x}$ \;
	\textbf{return:} $\beta_t$, $\bm{\hat{\sigma}}$.
	\caption{The Nishimori-Bethe relation for node classification}
	\label{alg:2}
\end{algorithm}

\subsection{The Nishimori temperature-based node classification algorithm}

For the purpose of node clustering though, the knowledge of $\beta_{\rm N}$ is a necessary prerequisite to obtain a precise estimate of the genuine node classes $\bm\sigma$. We indeed show next that a powerful estimator of $\bm\sigma$ is obtained directly from the signs of the entries of the eigenvector $\bm x$ of the Bethe-Hessian matrix $H_{\beta_{\rm N},\tilde J}$ (so $\beta_{\rm N}$ needs be known) associated to its smallest amplitude eigenvalue (which we now know is close to zero).

To this end, let us first consider $\bm{y}$, the eigenvector associated to the smallest eigenvalue of $H_{\beta_{\rm N}, J}$. Denote with $A \in \{0,1\}^{n \times n}$ the symmetric adjacency matrix of $\mathcal{G}$, defined by $A_{ij} = 1$ if $(ij) \in \mathcal{E}$, and $A_{ij}=0$ otherwise, and let $D \in \mathbb{N}^{n\times n}$ be the diagonal degree matrix $D = {\rm diag}(A\bm{1}_n)$. Then, applying Property~\ref{prop:th}, one easily obtains that
\begin{align}
\mathbb{E}\left[H_{\beta_{\rm N},J}\right] = I_n + \mathbb{E}\left[\frac{{\rm th}(\beta J_{ij})}{1 - {\rm th}^2(\beta J_{ij})}\right]\left(D - A\right).
\label{eq:exp_H}
\end{align}
From a straightforward calculation (see Proposition~1 of \citep{von2007tutorial}), the vector $\bm{1}_n$ is the eigenvector of $\mathbb{E}[H_{\beta_{\rm N}, J}]$ associated to its eigenvalue of smallest amplitude. As a consequence, from the relation between $\bm x$ and $\bm y$ (or equivalently between $\tilde J$ and $J$) in the previous section, the vector $\bm{\sigma}$ is the eigenvector of $\mathbb{E}[H_{\beta_N,\tilde{J}}]$ associated with its eigenvalue of smallest amplitude. Consequently, the eigenvector with zero eigenvalue of $H_{\beta_{\rm N}, \tilde{J}}$ is a close approximation\footnote{Rigorously, it is not so straightforward to move from $\mathbb{E}[H_{\beta_{\rm N}, J}]$ to $H_{\beta_{\rm N}, J}$. In \citep{dall2019revisiting}, a similar setting is considered in which the eigenvector $\bm{x}$ is studied in depth. The article argues that the relation indeed holds.} of $\bm{\sigma}$.

\medskip

This conclusion immediately translates into Algorithm~\ref{alg:2}, a numerical method to infer the genuine node classification $\bm{\sigma}$. Further detail on the practical implementation of this algorithm are provided in Appendix~\ref{app:algo}.

\medskip

Having established a ``Nishimori-optimal'' version of the Bethe Hessian-based spectral clustering for node classification, the next section discusses the relation between the proposed algorithm and other commonly used kernel matrices in the spectral clustering literature.

\begin{figure}[!t]
	\centering
	\includegraphics[width=\columnwidth]{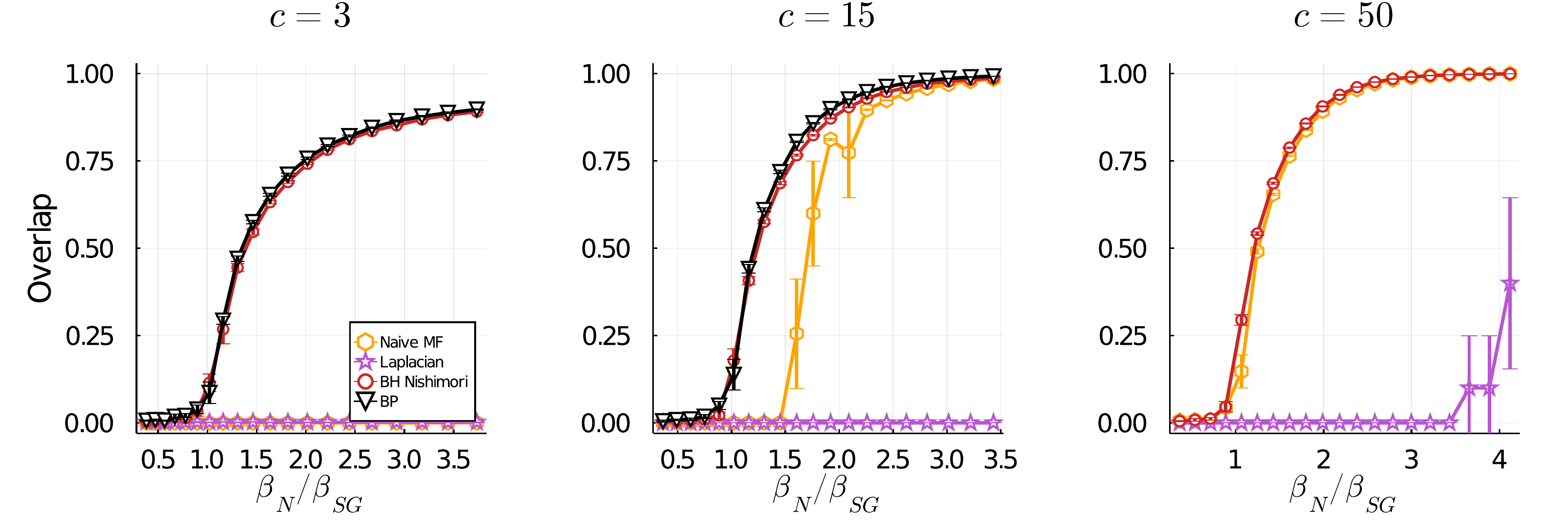}
	\caption{Overlap performance as a function of $\beta_{\rm N}/\beta_{\rm SG}$ and three different values of the expected average degree, $c$. For $\beta_{\rm N} < \beta_{\rm SG}$ inference is asymptotically unfeasible. Two classes of equal size are considered and the entries of $J$ are generated independently according to a Gaussian with mean $J_0\sigma_i\sigma_j$. In the examples, $n = 30.000$ and n average is taken over $10$ simulations.}
	\label{fig:perf_synth}
\end{figure}

\subsection{Relation to other spectral methods}
\label{sec:ml.relation}

In the following, we use the \emph{overlap}
\begin{align}
{\rm Overlap} =  \left|2\left(\frac{1}{n}\sum_{i = 1}^n \delta_{\sigma_i,\hat{\sigma}_i} - \frac{1}{2}\right)\right|.
\end{align}
as a measure of comparison of the inference performance of various node classification algorithms, where $\hat\sigma_i$ is the estimated label of node $i$. The overlap ranges from $0$ (random assignment) to $1$ (perfect assignment). Figure~\ref{fig:perf_synth} compares the overlap achieved by Algorithm~\ref{alg:2} versus the na\"ive mean field approach, consisting in estimating the labels from the dominant eigenvector of $\tilde{J}$, and versus the popular legacy spectral clustering algorithm based on the weighted graph Laplacian matrix $L = \bar{D} - \tilde{J}$, where $\bar{D} = {\rm diag}(|\tilde{J}|\bm{1}_n)$ \citep{kunegis2010spectral}.\footnote{Here  $|\cdot|$ is the entry-wise absolute value.} The figure browses several values of $\beta_{\rm N}$ (the larger $\beta_{\rm N}$, the easier the detection problem) and of the average degree $c$. For $c = 3, 15$ the output of the asymptotically optimal belief propagation (BP) algorithm is further shown, evidencing that Algorithm~\ref{alg:2} achieves an almost optimal performance. Due
to its computational complexity, we chose not to run belief propagation for $c = 50$, but we
expect to observe a similar result to the one obtained for $c = 3, 15$.
Before discussing the achieved results, let us first justify our comparison choice by recalling the rationale behind the Laplacian and na\"ive mean-field approaches.

\subsubsection{The weighted Laplacian matrix}

A very classical spectral clustering method in weighted graphs (dating back from the earliest works on the subject \citep{von2007tutorial}) exploits the \emph{weighted Laplacian} matrix $L = \bar{D} - \tilde{J}$, where $\bar{D} = {\rm diag}(|\tilde{J}|\bm{1}_n)$. As shown in \citep{von2007tutorial,kunegis2010spectral}, the eigenvector attached to the smallest eigenvalue of $L$ provides a (discrete to continuous) relaxed solution of the NP-hard optimization \emph{signed ratio-cut} graph clustering problem. The idea underlying the signed ratio-cut procedure consists in inferring the label assignments $\bm\sigma$ by maximizing the number of edges with positive weights connecting nodes in the same community, while minimizing the number of edges with negative weights connecting nodes in opposite communities: this is however a discrete optimization problem, a continuous relaxation of which coincides with a minimal eigenvector problem for $L$.

A particularly immediate and best understood scenario is the case of \emph{signed graphs}, in which the entries of $\tilde{J}$ assume values in $\pm 1$. For this class of graphs, an explicit relation between the matrices $L$ and $H_{\beta_{\rm N}, \tilde{J}}$ arises in the limit of trivial clustering, \emph{i.e.}, as $\beta_{\rm N} \to \infty$. For signed graphs, a slightly different definition of $H_{\beta, \tilde{J}}$ than \eqref{eq:BH_th} is most appropriate:
\begin{align}
H^{\rm signed}_{\beta,\tilde{J}} = (1 - {\rm th}^2(\beta))I_n + {\rm th}^2(\beta) D - {\rm th}(\beta)\tilde{J}.
\label{eq:BH_signed}
\end{align}
It is straightforward to notice that the signed and unsigned versions of the Bethe-Hessian matrix share the same set of eigenvectors \emph{on a signed graph} while their eigenvalues only differ by a multiplicative constant. One then immediately finds that $\lim_{\beta_{\rm N} \to \infty} H_{\beta_{\rm N},\tilde{J}}^{\rm signed} = L$. The signed Laplacian may then be seen as the \emph{zero temperature limit} of the Bethe-Hessian matrix. From a Bayesian inference standpoint \eqref{eq:inference_Nishi}, $H_{\beta_{\rm N},\tilde{J}}^{\rm signed}$ is a linear approximation of the exact inference problem, while $L$ is only an approximation for the \emph{maximum a posteriori} probability problem.\footnote{Taking the limit $\beta_{\rm N} \to \infty$ in Equation~\eqref{eq:inference_Nishi} is equivalent to looking for the maximum a posteriori solution.} Far from the limit of trivial recovery, our proposed Bethe-Hessian matrix-based method is thus expected to accomplish better inference performance when compared to the weighted Laplacian approach. This is indeed confirmed by Figure~\ref{fig:perf_synth}, which evidences a striking performance gap between both methods. The reconstruction performance achieved through the matrix $L$ is in particular severely compromised in the sparse regime in which $\tilde{J}$ only has $O_n(n)$ non-zero entries. 

\subsubsection{The na\"ive mean field approach}

The ``Nishimori Bethe-Hessian'' matrix is built from the Bethe approximation of the Bayes optimal problem formulation. We now show that a similar approximation procedure could have been performed using a na\"ive mean field approximation instead. This leads to a different -- much less efficient as we will see -- spectral clustering algorithm. Recalling the procedure of Section~\ref{sec:variational}, we define the na\"ive mean field free energy from the probability distribution
\begin{align}
p_{\bm{m}}(\bm{s}) = \prod_{i \in\mathcal{V}} \frac{1 + m_is_i}{2},
\label{eq:MF_distr}
\end{align}
where $m_i$ is the average of $s_i$ over the distribution~\eqref{eq:MF_distr}. The associated variational free energy reads
\begin{align*}
\tilde{F}_{\tilde{J},\beta}^{\rm MF}(\bm{m}) = -\sum_{(ij) \in \mathcal{E}} \beta \tilde{J}_{ij}m_im_j + \sum_{i \in \mathcal{V}}\sum_{s_i} \frac{1+ m_is_i}{2}~{\rm log}\left(\frac{1+m_is_i}{2}\right).
\end{align*}
Computing the gradient of $\tilde{F}^{\rm MF}_{\tilde{J},\beta}(\bm{m})$, one finds that, also in this case, the paramagnetic point $\bm{m} = \bm{0}$ is an extreme. Computing the Hessian of the free energy at the paramagnetic point leads instead to
\begin{align}
H_{\beta, \tilde{J}}^{\rm MF} = I_n - \beta \tilde{J}.
\end{align}
As a consequence, despite the presence of $\beta$ in the formulation of $H_{\beta, \tilde{J}}^{\rm MF}$, the eigenvectors of $H_{\beta,\tilde{J}}^{\rm MF}$ are simply the eigenvectors of $\tilde{J}$ so that, in this case, $\beta$ plays no role. Under the sparse regime, where $c = O_n(1)$, using the eigenvector associated to the smallest (resp., largest) eigenvalues of $H_{\beta,\tilde{J}}^{\rm MF}$ (resp., $\tilde{J}$) as an estimator for $\bm\sigma$ does not allow to make non-trivial reconstruction as soon as theoretically possible, \emph{i.e.}, whenever $\beta_{\rm N} > \beta_{\rm SG} > \beta_{\rm F}$: in this case indeed, the asymptotic spectrum of $\tilde{J}$ is unbounded and no isolated eigenvalue of $H_{\beta,\tilde{J}}^{\rm MF}$ is to be found. This explains the poor performance depicted in Figure~\ref{fig:perf_synth}  for small average degrees $c$. On the opposite, as already observed in Section~\ref{sec:main.main}, for sufficiently large degrees $c$, the na\"ive mean field approximation essentially yields the same result as the Bethe approximation. 

\begin{figure}[!t]
	\centering
	\includegraphics[width=\columnwidth]{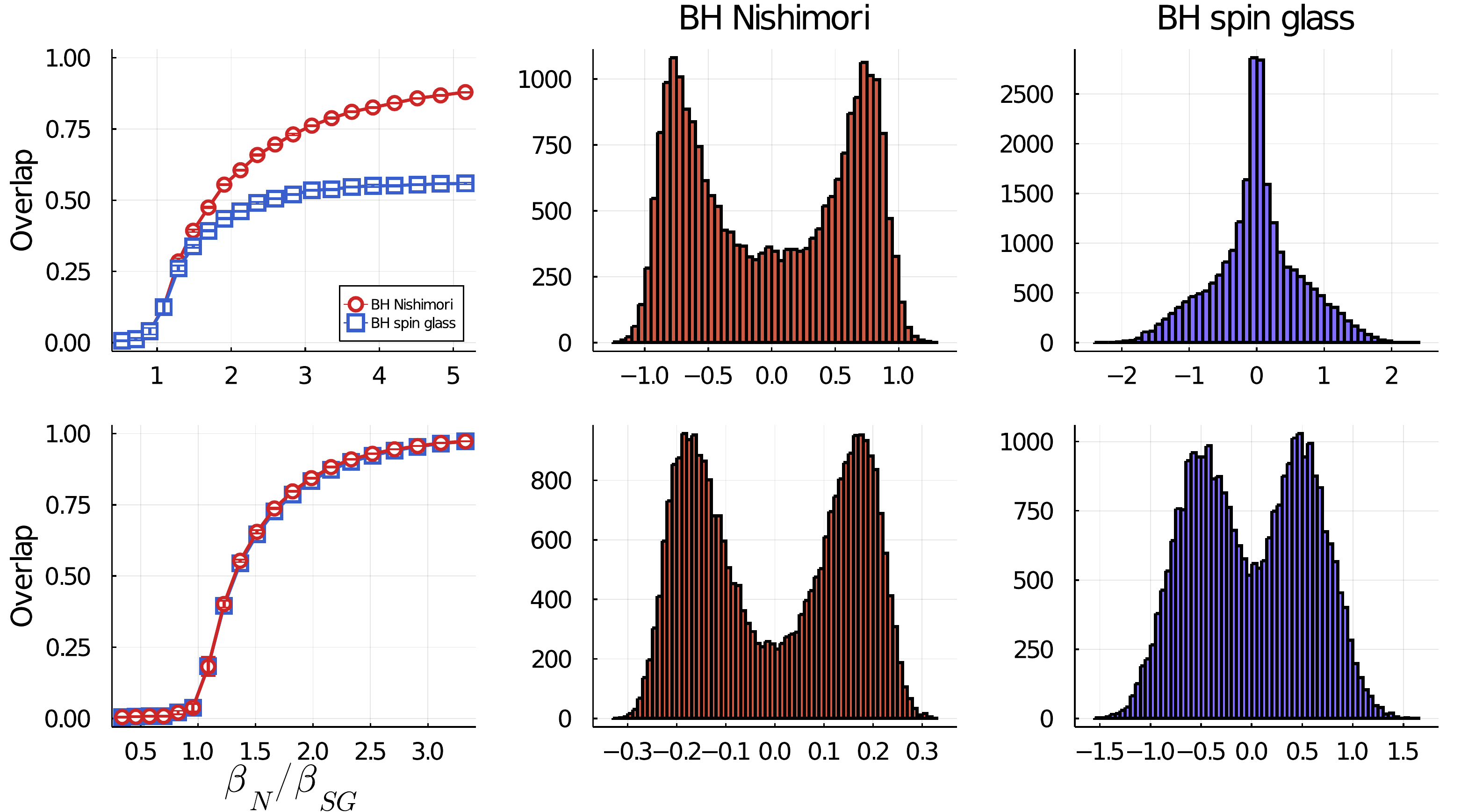}
	\caption{\textbf{First row}: random graphs with an underlying power law degree distribution. \textbf{Second row}: Erd\H{o}s R\'enyi random graphs.
	\textbf{First column}: overlap performance obtained exploiting the eigenvector associated to the smallest eigenvalue of $H_{\beta_{\rm N},\tilde{J}}$ (red circles) $H_{\beta_{\rm SG}, \tilde{J}}$ (blue squares). The entries of $\tilde{J}$ are distributed according to a Gaussian measure as in Equation~\eqref{eq:Nish_distr_inf}. Averages are taken over $10$ realizations. \textbf{Second column}: histogram of the entries of the informative eigenvector of $H_{\beta_{\rm N}, \tilde{J}}$ for $\beta_{\rm N}/\beta_{\rm SG} \approx 3.6$ in the first plot. \textbf{Third column}: histogram of the entries of the informative eigenvector of $H_{\beta_{\rm SG}, \tilde{J}}$ for the same configuration as the second plot. \textbf{For all plots}: the graphs have $n = 30~000$ nodes and expected average degree $c = 10$.}
	\label{fig:degree}
\end{figure}

\subsubsection{The ``spin glass Bethe-Hessian''}

We conclude this section by presenting an alternative use of the Bethe-Hessian matrix, inspired by the work of \citep{saade2016clustering}, that we name here the \emph{spin glass Bethe-Hessian}. Algorithm~\ref{alg:2} represents an optimal relaxation of the Bayes optimal solution, capable of performing better than random inference as soon as theoretically possible. The parametrization $\beta = \beta_{\rm N}$ is not the only possible choice of $\beta$ able to reach this threshold. It was indeed shown, under different settings, in \citep{saade2014spectral, saade2016clustering, dall2020community, shi2018weighted} that choosing the temperature $\beta = \beta_{\rm SG}$ allows one also to achieve non-trivial clustering as soon as theoretically possible. 

The value $\beta_{\rm SG}$, unlike $\beta_{\rm N}$, can be easily estimated from the matrix $\tilde{J}$ solving $c\mathbb{E}[{\rm th}^2(\beta_{\rm SG} \tilde{J}_{ij}\sigma_i\sigma_j)] = c\mathbb{E}[{\rm th}^2(\beta_{\rm SG}\tilde{J}_{ij})] = 1$. However, it was proved in \citep{dall2019revisiting} that for community detection in realistic \emph{heterogeneous} (thus not Erd\H{o}s-R\'enyi-like) graphs, this may be a quite suboptimal choice in terms of the raw (say, overlap) classification performance. The main difference between the \emph{spin glass Bethe-Hessian} and the \emph{Nishimori Bethe-Hessian} is thus observed when the underlying graph is not of an Erd\H{o}s-R\'enyi type. This can be understood by a closer inspection of Equation~\eqref{eq:exp_H}, which shows that the vector $\bm{\sigma}$ is an approximate eigenvector of $H_{\beta_{\rm N},\tilde{J}}$ for \emph{any} underlying degree distribution of the graph. This would not be true in general for any other value of $\beta \neq \beta_{\rm N}$, hence in particular not for $\beta_{\rm SG}$. 

\medskip

As a visual confirmation, Figure~\ref{fig:degree} displays the overlap performance and the histograms of the entries of the informative eigenvector of $H_{\beta_{\rm N}, \tilde{J}}$ versus $H_{\beta_{\rm SG}, \tilde{J}}$ for a matrix $\tilde{J}$ generated according to Equation~\eqref{eq:Nish_distr_inf}, considering on the top row graphs with an underlying power-law degree distribution (this thus goes beyond the assumption of the present article, yet is typical of real-world graph models \citep{barabasi1999emergence}) and on the bottom row Erd\H{o}s-R\'enyi graphs. The loss in precision of the \emph{spin glass Bethe-Hessian} is best understood by comparing the two histograms which evidence that, unlike $H_{\beta_{\rm N},\tilde{J}}$, the underlying node classes seen by $H_{\beta_{\rm SG},\tilde{J}}$ is much spoiled by the heterogeneous degree distribution. This is also observed to some extent on Erd\H{o}s R\'enyi graphs, but here the performance achieved by $H_{\beta_{\rm SG}, \tilde{J}}$ is essentially the same as the one obtained with $H_{\beta_{\rm N},\tilde{J}}$.

The use of $H_{\beta_{\rm N}, \tilde{J}}$ should thus be privileged when the input weighted graph $\mathcal G$ may be far from an Erd\H{o}s-R\'enyi random graph generation, such as in the case of a real-world weighted social graph. Besides, one can envision to extend Algorithm~\ref{alg:2} beyond two-class node clustering, as proposed in \citep{dall2020unified}, where the authors show that the proper parametrization of the Bethe-Hessian matrix (specifically using multiple rather than a single value for $\beta$) brings a decisive advantage on real datasets.

On the opposite, if the input graph is of the Erd\H{o}s-R\'enyi type, the performances of both algorithms are observed to be similar, with a slight computational as numerical stability  advantage for $H_{\beta_{\rm SG}, \tilde{J}}$. We nonetheless underline that the estimation of $\beta_{\rm N}$ may be of independent interest: if one uses the solution of spectral clustering as the initialization to an algorithm seeking the actual Bayes optimal solution, then the initialization provided by $H_{\beta_{\rm SG},\tilde{J}}$ would likely be of good quality, although $\beta_{\rm N}$ would still remain unknown.

\subsection{Application to real data classification}
\label{sec:application_realdata}
We complete the article by a robustness test of our proposed algorithm under a real-world machine learning classification problem. Specifically, we consider a \emph{sparse (and thus cost-efficient) version of} the problem of \emph{correlation clustering} such as met in image classification and show how Algorithm~\ref{alg:2} can be adopted to accomplish this task with higher performance than with competing spectral methods of the literature. 

\medskip

Let $\{\bm{z}_i\}_{i = 1,\dots, n}$ be an $n$-vector dataset with $\bm{z}_i \in \mathbb{R}^p$. These vectors represent discriminating \emph{features} of some two-class data (say images) to be clustered in a fully unsupervised manner. In typical modern machine learning, $p$ is of the order of a few thousands for images and a few hundreds for natural language text representations, and it is not rare to try and classify up to millions of data vectors $\bm{z}_i$. 

The most elementary unsupervised machine learning classification approach consists in running the popular \emph{k-means} algorithm in the ambient $p$-dimensional feature space. K-means is however known to fail for large $p$ \citep{kriegel2009clustering} and is ruled out as soon as $p$ exceeds the order of a few tens. A classical workaround is to \emph{embed} the feature vectors $\bm{z}_i$ in a lower dimensional space on which to run k-means clustering. The most popular embedding exploits a \emph{spectral} approach: one starts by defining a kernel matrix $K(\{\bm{z}\}) \in \mathbb{R}^{n \times n}$, the entry $K_{ij}(\{\bm{z}\})$ of which evaluates some \emph{affinity metric} between $\bm{z}_i$ and $\bm{z}_j$; running a principal component analysis on $K(\{\bm{z}\})$, one then extracts a collection of eigenvectors $\bm{x}_1,\ldots,\bm{x}_\ell$ for some $\ell$ of the order of the presumed number of classes; the rows $\tilde{\bm{x}}_1,\ldots,\tilde{\bm{x}}_n\in\mathbb R^\ell$ of the resulting ``tall'' matrix $\bm{X}=[\bm{x}_1,\ldots,\bm{x}_\ell]\in\mathbb R^{n\times \ell}$ form the embedding of the original features from $\mathbb R^p$ into $\mathbb R^\ell$ over which k-means clustering is finally run. A popular affinity function is merely the correlation $K_{ij}(\{\bm{z}\}) = \bm{z}_i^T\bm{z}_j$, which we shall consider here.\footnote{Other choices exist, such as the more popular \emph{heat} kernel $K_{ij}(\{\bm{z}\}) = \exp(-\|\bm{z}_i-\bm{z}_j\|^2/2\nu^2)$ for some $\nu>0$. 
}

\medskip

For large dimensional datasets though (\emph{i.e.}, for $p,n$ beyond a few thousands), the $O(pn^2)$ cost of building $K(\{\bm{z}\})$ added to the (at least) $O(n^2)$ cost of the principal component analysis step makes spectral clustering hardly achievable on a modern home computer. To drastically decrease the computational complexity one may proceed to a two-level sparsification as recently proposed in \citep{zarrouk2020performance,couillet2021twoway}: by randomly discarding elements of the $p$-dimensional features $\bm{z}_i$ and by randomly dropping a number of evaluations of the correlations $\bm{z}_i^T\bm{z}_j$. This operation of course impedes the clustering performance, but, as surprisingly proved in \citep{zarrouk2020performance,couillet2021twoway} under a ``still rather dense graph'' regime, the performance loss is negligible for a wide range of sparsity levels. To this end, let $S \in \{0,1\}^{n\times p}$ and $M \in \{0,1\}^{n \times n}$ (symmetric) be Bernoulli masks with parameters $\sqrt{\kappa/p}$ and $c/n$,\footnote{The choice of $c$ is not a coincidence: $M$ will enforce an average node degree of $c$ to the resulting graph.} respectively. The resulting sparsified kernel matrix then becomes
\begin{align}
\tilde{J} &= K(\{\tilde{\bm{x}}\}) \circ M,\quad \text{where} \quad \tilde{x}_{i,l} = x_i S_{i,l}.
\end{align}
\emph{i.e.}, each entry of each of the feature vectors $\bm z_i$ is kept only with probability $\sqrt{\kappa/p}$, while each measurement $K_{ij}(\{\tilde{\bm{z}}\})$ is only performed with probability $c/n$. The computational complexity to build $\tilde{J}$ is thus scaled down to $O(\kappa c n)$, \emph{i.e.}, to linear time complexity with respect to the size of the original dataset (which is the best one can hope for without completely dropping part some of the data $\bm{z}_i$). 

As a major consequence of the sparsification procedure, the non-zero entries of $\tilde{J}$ can be considered asymptotically independent due to the asymptotic absence of short loops in the underlying sparse Erd\H{o}s-R\'enyi graph. As a result, Equation~\eqref{eq:Nish_distr_inf} provides a good approximation for the generative model of $\tilde{J}$ and for a two-class correlation clustering problem, Algorithm~\ref{alg:2} can be efficiently used on the matrix $\tilde J$.

\medskip

We thus practically tested Algorithm~\ref{alg:2} against the na\"ive mean field approach which in this setting happens to coincide with the algorithm proposed in \citep{zarrouk2020performance} when applied to the $\tilde{\bm{z}}_i$ vectors,
and against the weighted Laplacian matrix approach. As a telling modern data classification context, we chose to cluster two classes of high-resolution extremely realistic images randomly produced by \emph{generative adversarial networks} (the now quite popular GANs) \citep{brock2018large}; the interest of using GAN images rather than real images lies in that GAN images can be produced ``on-the-fly'' and in arbitrary numbers.
\begin{figure}
	\centering
	\includegraphics[width=\columnwidth]{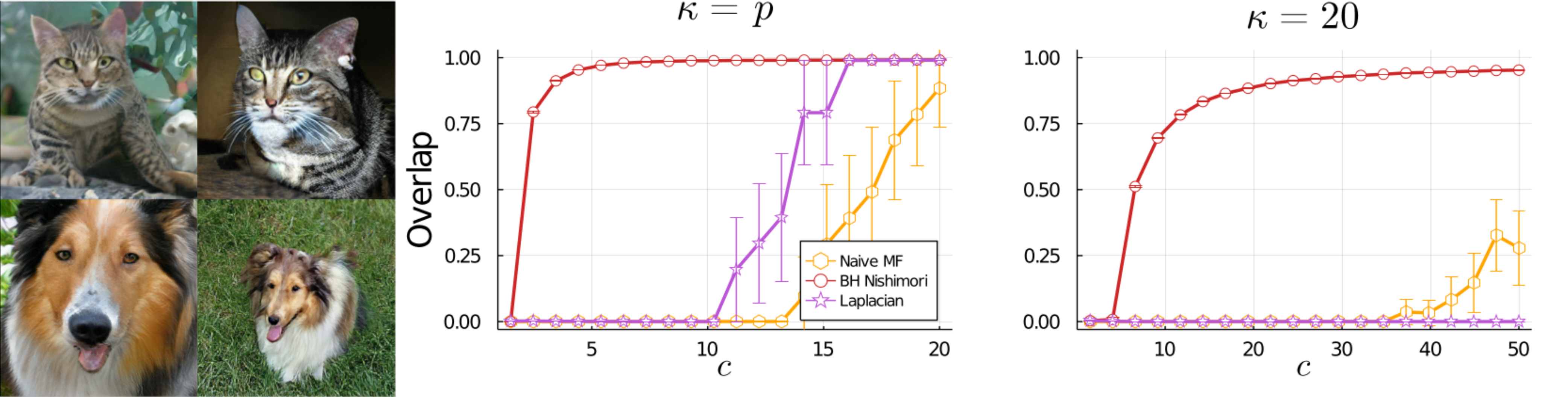}
	\caption{\textbf{Left plot}: example of random generations of GAN images representing \texttt{collie dogs} and \texttt{tabby cats} used for the experiment. \textbf{Right plots}: overlap classification performance of $40\,000$ GAN images, as a function of the expected average underlying graph degree $c$. Here, we consider  $K_{ij}(\{\bm{x}\}) = \frac{1}{p}\bm{x}_i^T\bm{x}_j$ and we take either $\kappa = p$: all features of the images are kept, or $\kappa = 20$: on average, only $\sqrt{\kappa/p}$ features (out of the original $p=512$) are used. Simulation performances are themselves averaged over $10$ realizations.}
	\label{fig:GAN}
\end{figure}

\medskip

Specifically, we considered $n = 40\,000$ images divided into two groups of equal size, representing collie dogs and tabby cats. A representative example of the input images generated by the GAN is given in Figure~\ref{fig:GAN}. For each of these images we extracted discriminating features using an off-the-shelf convolutional neural network (VGG) which produces $p = 512$-dimensional feature vectors $\bm{z}_i$.\footnote{The $p=512$ figure is on the low-hand of typical image vector representations: this number today may rise to $4k$ or even to $20k$ when much more than two classes of images are to be classified.} We then measured the overlap performance as a function of the average node degree $c$ of the ensuing graph and for different values of $\kappa$. The results are reported in Figure~\ref{fig:GAN} which strikingly evidences that Algorithm~\ref{alg:2} can achieve almost perfect reconstruction already for $c = 5$ when the feature vectors $\bm z_i$ are not sparsified ($\kappa = p$): so, in clearer terms, out of the $40k\times 40k = 1.6\cdot 10^9$ correlations needed to evaluate the full $K(\{\bm{z}\})$ matrix, only $\approx 6\times 40k=2.4\cdot 10^5$ is enough to achieve almost optimal performance, thus corresponding to a striking $10^4$-fold gain in complexity for a rather marginal performance loss!

Figure~\ref{fig:GAN} also reports that the performance of the na\"ive mean-field and weighted Laplacian matrix approaches, currently the legacy methods in the literature, severely suffer in the low-$c$ end. These observations perfectly adhere with the conclusions drawn so far in the article and thus turns our up-to-here formal \emph{Nishimori-optimized} algorithm into a concrete powerful method for cost-efficient classification of large dimensional datasets.

\section{Conclusion}
\label{sec:conclusion}

The central contribution of the article is of a theoretical nature and aims at introducing an elegant explicit relation between the Bethe-Hessian matrix and the Nishimori temperature. Yet, beyond this statistical physics endeavor, which will surely find further independent theoretical interests, the result finds fundamental direct applications to Bayesian statistical inference; this is strikingly evidenced by the image clustering application devised in Section~\ref{sec:application_realdata}. Specifically, one may anticipate an important impact in more involved applications than those considered in this article, such as in \emph{restricted Boltzmann machines} (RBM) whose goal is to learn a generative model from a set of examples \citep{ackley1985learning}: the Bethe approximation has recently been adopted to study the RBM from a Bayesian perspective \citep{huang2016unsupervised} so that one may envision that the explicit relation between the Bethe free energy and the Bayes optimal (Nishimori) condition presented in this article would lead to a better understanding and improvement of state-of-the-art algorithms. Similarly, the Bethe and TAP approximations have recently been exploited to devise efficient spectral algorithms for phase retrieval, based on statistical physics intuitions similar to the ones detailed in this article  \citep{luo2019optimal, ma2021spectral, maillard2020construction}. The extension of our results to this more involved setting is a promising line of exploration.

\medskip 

On the side of complexity reduction, exploiting high levels of sparsification of data measurements, we showed that our proposed algorithm is capable of accomplishing high quality \emph{unsupervised} classification on very large datasets. This result is all the more fundamental that future machine learning data treatment will call for increasingly larger datasets which cannot be possibly manually labelled and for which unsupervised (or possibly semi-supervised) approaches must be adopted.\footnote{A configuration which, in passing, even modern so-called \emph{deep} neural networks struggle to correctly handle.} As a downside though, the generative model we considered for the data affinity (kernel) matrix takes the strong assumption that its entries are drawn from the same probability distribution and only the average (and not the variance, or the distribution itself) embeds information on the node labels.  This setting might be too simplistic on generic real data that would require more realistic probability distributions for the generative model of the kernel matrix, considering, for instance, asymmetrical \citep{saade2016clustering}, multi-cluster \citep{shi2018weighted} or multi-dimensional distributions.

Possibly most importantly, we worked here under the assumptions that the edges maintained in the sparsified graph are drawn independently at random. When dealing with actual kernel matrices, this cost-efficient measure is quite suboptimal: in \citep{liao2020sparse}, a more efficient sparsification procedure is used which maintains the entries of $K(\{\bm{z}\})$ of largest amplitude. In \citep{liao2020sparse}, this comes at the cost of computing all the entries of $K(\{\bm{z}\})$ but, surely, a more efficient nearest neighbors-type procedure could be implemented as a good performance-complexity compromise \citep{muja2009fast}. Yet, in this setting, although stronger sparsity levels can surely be achieved for the same performance, the key independence property of the entries of $K(\{\bm{z}\})$ which we exploited here can no longer be assumed, so that one needs to carefully handle the hard problem of dependencies. There lies the main objectives of our follow-up investigations.

\section*{Acknowledgements}

RC’s work is supported by the MIAI LargeDATA Chair at University Grenoble-Alpes and the GIPSA-HUAWEI Labs project Lardist. NT's work is partly supported by the French National Research Agency in the framework of the "Investissements d’avenir” program (ANR-15-IDEX-02) and the LabEx PERSYVAL (ANR-11-LABX-0025-01).
The authors thank Mohamed El Amine Seddik for sharing the codes to produce the experiments on GAN images.


\setcitestyle{numbers}
\bibliographystyle{unsrt}
\nocite{*}
\bibliography{nishimori}

\newpage

\appendix

\section{An explicit expression for the matrix $F({g})$}
\label{app:F_matrix}

We here provide one of the possible explicit expressions that the matrix $F(\bm{g})$ can have. In particular, this is the  expression used in our simulations. Let us recall the definition of the matrix $F(\bm{g})$. 

Let $\bm{g} \in\mathbb{R}^{2|\mathcal{E}|}$ be an eigenvector of the matrix $B$ with weight vector $\bm{\omega} \in \mathbb{R}^{2|\mathcal{E}|}$. Let $\lambda$ be the eigenvalue corresponding to $\bm{g}$, with $|\lambda| \geq 1$. The matrix $F(\bm{g})$ is \emph{any} matrix satisfying the relation
\begin{align}
\left[F(\bm{g})\bm{\psi}(\bm{g})\right]_i = \sum_{j \in\partial i} \omega_{ij}^3 g_{ij},
\label{eq:def_F_2}
\end{align}
where we recall that 
\begin{align*}
\psi_i(\bm{g}) = \sum_{j \in \partial i} \omega_{ij}g_{ij}.
\end{align*}
A possible definition of the matrix $F(\bm{g})$ is to consider a diagonal matrix, satisfying
\begin{align*}
F_{ij}(\bm{g}) = \delta_{ij}\frac{\sum_{j \in \partial i} \omega_{ij}^3 g_{ij}}{\sum_{j \in \partial i} \omega_{ij}g_{ij}}.
\end{align*}
This matrix depends however explicitly on $\bm{g}$. We here describe an alternative expression in which the dependence on $\bm{g}$ is manifested only through $\lambda$. More explicitly, the  following relation holds
\begin{align*}
\lambda g_{ij} = (B\bm{g})_{ij} = \psi_j(\bm{g}) - \omega_{ij}g_{ji}.
\end{align*}
Considering the same equation for $g_{ji}$, we can easily write the following system
\begin{align*}
\begin{pmatrix}
\lambda & \omega_{ij} \\ 
\omega_{ij} & \lambda
\end{pmatrix}\begin{pmatrix}
g_{ij} \\
g_{ji}
\end{pmatrix} = \begin{pmatrix}
\psi_j(\bm{g}) \\
\psi_i(\bm{g})
\end{pmatrix}.
\end{align*}
For $|\lambda|  \geq 1$ and $|\omega_{ij}| < 1$, the matrix on the left hand-side can be  inverted, leading to the following relation
\begin{align}
g_{ij} = \frac{\lambda \psi_j - \omega_{ij} \psi_i}{\lambda^2 - \omega_{ij}^2}.
\label{eq:g_ij}
\end{align}
Plugging Equation~\eqref{eq:g_ij} into Equation~\eqref{eq:def_F_2}, the following expression of $F(\bm{g}) \equiv F(\lambda)$ can be obtained:
\begin{align*}
F_{ij}(\lambda) = -\delta_{ij}\sum_{k \in \partial i}\frac{\omega_{ij}^4}{\lambda^2-\omega_{ij}^2} + \frac{\lambda\omega_{ij}^3}{\lambda^2 - \omega_{ij}^2}.
\end{align*}
This expression of the matrix $F(\bm{g})\equiv F(\lambda)$ is the one considered in our simulations.

\section{Algorithm implementation}
\label{app:algo}

In this appendix we discuss more extensively some details concerning a practical and efficient implementation of Algorithm~\ref{alg:2}. For reference, our codes are available at \href{https://github.com/lorenzodallamico/NishimoriBetheHessian}{github.com/lorenzodallamico/NishimoriBetheHessian}. We now proceed to a detailed analysis of each step of Algorithm~\ref{alg:2}.

\medskip

The first step of Algorithm~\ref{alg:2} consists in the following operation:
\begin{align*}
    \forall~(ij) \in \mathcal{E}~:~ \tilde{J}_{ij} = \tilde{J}_{ij} - \frac{1}{2|\mathcal{E}|}\bm{1}_n^T\tilde{J}\bm{1}_n.
\end{align*}
The rationale of this operation is to consider an input matrix $\tilde{J}$ as close as possible to a realization of the distribution of Equation~\eqref{eq:Nish_distr_inf} that satisfy, for two classes of equal size\footnote{Note that  the inference problem of Equation~\eqref{eq:inference_Nishi} does not make any assumption on the respective sizes of the classes that can therefore be arbitrary. In the case of asymmetric classes, however, the term $\mathbb{E}[J_{ij}] \neq 0$ depends on the sizes of the two classes. In order to do the proper shift, one would then need additional information on the class sizes.}, the condition $\mathbb{E}[\tilde{J}_{ij}] = 0$. By shifting the  empirical average of $\tilde{J}_{ij}$ to zero for the input of Algorithm~\ref{alg:2}, we are willing to reproduce this property. Note that \emph{only the non-zero entries of $\tilde{J}$ are shifted}, while for all the $(ij) \notin \mathcal{E}$ the $\tilde{J}_{ij} = 0$. 

\medskip

Once a proper input matrix $\tilde{J}$ is obtained, the value of $\beta_{\rm SG}$ and then the smallest eigenvalue of $H_{\beta_{\rm SG}, \tilde{J}}$ are computed: if the latter is positive, one cannot proceed any further to the computation of $\hat{\beta}_{\rm N}$ and the algorithm is stopped. In the spirit of \emph{correlation clustering}, discussed in Section~\ref{sec:ml}, the condition $\gamma_{\rm min}(H_{\beta_{\rm SG},\tilde{J}}) < 0$ imposes the minimal average degree to perform non-trivial reconstruction.\footnote{The detectability condition we recall to imposed by $\beta_{\rm N} > \beta_{\rm SG}$. While $\beta_{\rm N}$ is independent of the average degree, $\beta_{\rm SG}$ is a decreasing function of the average degree, as it can be easily obtained from its definition in Equation~\eqref{eq:def_transitions_SG}.}

\medskip

At this point, we get to the core of Algorithm~\ref{alg:2} that consists in the computation of $\hat{\beta}_{\rm N}$. The first thing to do is to determine if $\mathcal{G}$ is a signed graph (with only $\pm J_0$ entries). If this is the case, the signed representation of $H_{\beta,\tilde{J}}$ introduced in Equation~\eqref{eq:BH_signed} should be adopted. We consider first this easier case. 

For notation convenience, we introduce $r = [{\rm th}(\beta J_0)]^{-1}$ ($r \geq 1$)  and define $H_{r, \tilde{J}} = (r^2-1)I_n+D-r\tilde{J}$. Furthermore, let $\bm{x}_{r}$ be the eigenvector of $H_{r,\tilde{J}}$ associated to its smallest eigenvalues. We look for $r$ so that $$\gamma_{\rm min}\left(H_{r,\tilde{J}}\right)~=~0.$$
In order to do so,  consider $r_t > \hat{r}_{\rm N} = [\rm{th}(\hat{\beta}_{\rm N}J_0)]^{-1}$. The following relation is true for any $r$,
\begin{align}
    \gamma_{\rm min}(H_{r,\tilde{J}}) \leq \bm{x}_{r_t}^TH_{r,\tilde{J}}\bm{x}_{r_t} = (r^2-1) + d_{r_t} - r\mathsf{j}_{r_t} := f_{r_t}(r),
    \label{eq:CF}
\end{align}
where $d_{r_t} = \bm{x}_{r_t}^TD\bm{x}_{r_t}$ and $\mathsf{j}_{r_t} = \bm{x}_{r_t}^T\tilde{J}\bm{x}_{r_t}$. Defining $r_{t+1}$ as the solution to $f_{r_t}(r_{t+1}) = 0$, one immediately obtains from Equation~\eqref{eq:CF} that $\gamma_{\rm min}(H_{r_{t+1},\tilde{J}}) < 0$. One can show \citep[Appendix F]{dall2020community} that $|r_{t+1} - \hat{r}_{\rm N}| < |r_t - \hat{r}_{\rm N}|$, hence, that at each iteration the value of $r_t$ approaches $\hat{r}_{\rm N}$. In practice, convergence is typically achieved in less than $10$ iterations. A good initialization is $r_0 = [{\rm th}(\beta_{\rm SG}J_0)]^{-1} > \hat{r}_{\rm N}$ (recall that $\beta_{\rm N} > \beta_{\rm SG}$), ensuring the algorithm convergence.

We now consider graphs with non-binary weights that introduce additional complications. The entries of $H_{\beta, \tilde{J}}$ grow exponentially, with $\beta$, making the eigenvalue computation potentially unstable. In order to work with a matrix with entries of order $1$, we introduce the following \emph{weighted regularized Laplacian} \citep{dall2020optimal}:
\begin{align}
L_{\beta, \tilde{J}}= I_n - \Lambda_{\beta, \tilde{J}}^{-1/2} \tilde{W}_{\beta, \tilde{J}} \Lambda_{\beta, \tilde{J}}^{-1/2}, 
\label{eq:L}
\end{align}
where 
\begin{align*}
\left(\tilde{W}_{\beta, \tilde{J}}\right)_{ij} = \frac{{\rm th}(\beta \tilde{J}_{ij})}{1-{\rm th}^2(\beta \tilde{J}_{ij})}; \quad 
\left(\Lambda_{\beta, \tilde{J}}\right)_{ij} = \delta_{ij}\sum_{k \in \partial i}\frac{{\rm th}^2(\beta \tilde{J}_{ik})}{1-{\rm th}^2(\beta \tilde{J}_{ik})}
\end{align*}
It is straightforward to see that if $H_{\hat{\beta}_{\rm N}, \tilde{J}}\bm{x} = 0$, then $L_{\hat{\beta}_{\rm N}, \tilde{J}}\bm{v} = 0$, where  $\bm{v} = \Lambda_{\beta,\tilde{J}}^{-1/2}\bm{x}$. 
The matrix $L_{\beta,\tilde{J}}$ hence allows to compute $\hat{\beta}_{\rm N}$ and $\bm{x}$ in a more efficient way, since it is more suited to eigenvalue computations. We can define in this case $f_{\beta_t}(\beta) = \bm{v}_t^TL_{\beta,\tilde{J}}\bm{v}_t$ and update $\beta_{t+1}$ as the solution to $f_{\beta_t}(\beta_{t+1}) = 0$.

In any case, for very large values of $\beta_{\rm N}$ (hence for very easy clustering problems) numerical instabilities may occur. In order to avoid this problem, we allow a ``maximal value" $\beta_{\rm th}$ for $\hat{\beta}_{\rm N}$ beyond which the algorithm is stopped. The main reason that allows us to do so is that if $\beta_{\rm N} > \beta_{\rm th}$ we are practically in a easy detection regime, for which the knowledge of the exact value of $\beta_{\rm N}$ is less relevant and can be otherwise achieved first estimating the labels $\bm{\hat{\sigma}}$ (the estimation of which will be very accurate) and then solving $\mathbb{E}[{\rm th}(\beta_{\rm N}\tilde{J}_{ij}\sigma_i\sigma_j)] = \mathbb{E}[{\rm th}^2(\beta_{\rm N}\tilde{J}_{ij})]$.
We empirically observed that a good stopping criterion is obtained imposing $\beta_{\rm th} \sim \sqrt{c}\beta_{\rm SG}$.

\end{document}